%% file: aaai2027.tex
\documentclass[letterpaper]{article} 
\usepackage[preprint]{aaai2027}  

\usepackage[hyphens]{url} 
\usepackage{graphicx}
\usepackage{natbib}
\usepackage{caption}
\usepackage{booktabs}
\input{00_package}
\usepackage{algorithm}
\usepackage{algorithmic}

\usepackage{newfloat}
\usepackage{listings}
\usepackage{tikz}

\usepackage{array}
\usepackage[table]{xcolor}
\usepackage{tabularx}

\newcommand{\dashunderline}[1]{%
  \tikz[baseline=(txt.base)]{%
    \node[inner sep=0pt, outer sep=0pt] (txt) {#1};%
    \draw[
      dash pattern=on 1.1pt off 0.8pt,
      line width=0.35pt
    ]
    ([yshift=-0.45ex]txt.south west) --
    ([yshift=-0.45ex]txt.south east);%
  }%
}

\definecolor{rankfirst}{RGB}{0,120,80}
\definecolor{ranksecond}{RGB}{30,95,160}
\definecolor{rankthird}{RGB}{200,95,35}

\definecolor{corecolbg}{gray}{0.975}
\definecolor{corefirstbg}{RGB}{221,240,230}
\definecolor{coresecondbg}{RGB}{226,236,248}
\definecolor{corethirdbg}{RGB}{249,234,222}

\newcommand{\firstplace}[1]{\textbf{#1}}
\newcommand{\secondplace}[1]{\underline{#1}}
\newcommand{\thirdplace}[1]{\dashunderline{#1}}

\newcommand{\corefirst}[1]{%
  \cellcolor{corefirstbg}\textcolor{rankfirst}{\textbf{#1}$^{\scriptscriptstyle 1}$}%
}
\newcommand{\coresecond}[1]{%
  \cellcolor{coresecondbg}\textcolor{ranksecond}{\underline{#1}$^{\scriptscriptstyle 2}$}%
}
\newcommand{\corethird}[1]{%
  \cellcolor{corethirdbg}\textcolor{rankthird}{\dashunderline{#1}$^{\scriptscriptstyle 3}$}%
}

\newcommand{\summarysecond}[1]{\underline{#1}}
\newcommand{\summarythird}[1]{\dashunderline{#1}}

\DeclareCaptionStyle{ruled}{labelfont=normalfont,labelsep=colon,strut=off} 
\floatstyle{ruled}
\newfloat{listing}{tb}{lst}{}
\floatname{listing}{Listing}
\title{CORE: In-Context Reconstruction for Unified Tabular Anomaly Detection}

\author{
    Yunfeng Zhao\textsuperscript{\rm 1},
    Qingfeng Chen\textsuperscript{\rm 1}\corresponding,
    Yue Tan\textsuperscript{\rm 2},
    Shiyuan Li\textsuperscript{\rm 2},
    Yili Wang\textsuperscript{\rm 3},
    Yixin Liu\textsuperscript{\rm 2}\corresponding,
    Shirui Pan\textsuperscript{\rm 2}
}
\affiliations{
    \textsuperscript{\rm 1}Guangxi University\\
    \textsuperscript{\rm 2}Griffith University\\
    \textsuperscript{\rm 3}Jilin University\\
    \{yunf.zhao, qingfeng\}@gxu.edu.cn,\\
    \{yue.tan, shiyuan.li, yixin.liu, s.pan\}@griffith.edu.au,\\
    wangyl21@mails.jlu.edu.cn
}

\begin{document}

\maketitle

\begin{abstract}
\input{sections/0_abs}
\end{abstract}

\section{Introduction}

\input{sections/1_intro}

\section{Related Work}
\input{sections/2_rw}
\section{Preliminaries}
\input{sections/3_pre}

\section{Methodology}
\input{sections/4_method}

\section{Experiments}
\input{sections/5_exp}

\section{Conclusion}
\input{sections/6_con}


\bibliography{aaai2027}

\clearpage

\appendix

\setcounter{secnumdepth}{2}
\setcounter{section}{0}
\renewcommand{\thesection}{\Alph{section}}
\renewcommand{\thesubsection}{\Alph{section}.\arabic{subsection}}
\renewcommand{\thesubsubsection}{\Alph{section}.\arabic{subsection}.\arabic{subsubsection}}
\section{Detailing Related Work}\label{app:rw}
\input{Appendices/sections/1_rw}

\section{Algorithm and Complexity}\label{app:algo}
\input{Appendices/sections/2_algo}

\section{Details of Experimental Setup}
\input{Appendices/sections/3_detail}

\section{Supplemental Experiments}
\input{Appendices/sections/4_SuppleExp}

\end{document}

%% file: 00_package.tex
\usepackage{amsfonts}       
\usepackage{nicefrac}       
\usepackage{microtype}      
\usepackage{xcolor}         
\usepackage{xspace}
\usepackage{amsmath}
\usepackage{amssymb}
\usepackage{array}
\usepackage{subfigure}
\usepackage{pifont}
\usepackage{multirow}
\usepackage{color, colortbl}
\definecolor{Gray}{gray}{0.9}

\definecolor{fgreen}{RGB}{177,207,149}
\definecolor{fred}{RGB}{234,179,138}

\definecolor{firstcolor}{RGB}{20,128,85}
\definecolor{secondcolor}{RGB}{20,104,168}
\definecolor{thirdcolor}{RGB}{236,84,20}

\newcommand{\ourmethod}{CORE\xspace}

%% file: sections/0_abs.tex
Tabular anomaly detection (TAD), which focuses on identifying abnormal samples that deviate from the majority in tabular data, has received growing attention. 
Recently, there has been an emerging trend towards unified TAD, which seeks to detect anomalies across different datasets using a single generalizable model. 
In unified TAD, aligning heterogeneous data remains challenging. While existing methods often rely on distance-based unified feature construction, they may obscure the semantics of the original features. 
Moreover, existing approaches typically formulate anomaly detection as a binary classification task, which may overlook diverse anomaly patterns from various datasets and be misled by unrepresentative synthetic anomalies. 
To address these challenges, we propose an in-\textbf{CO}ntext \textbf{RE}construction approach for unified TAD (\ourmethod for short). It introduces a decorrelated feature alignment module to directly align heterogeneous features into a unified representation space, which retains their semantic information. 
Meanwhile, \ourmethod formulates unified TAD as an in-context reconstruction problem, eliminating the need for labeled or synthesized anomalies. Specifically, the in-context reconstruction module reconstructs each sample by leveraging contextual normal samples to capture dataset-specific distributions, such that reconstruction errors reflect its deviation from normality, facilitating unified TAD on arbitrary unseen datasets. 
Extensive experiments on 34 datasets from diverse domains demonstrate the superior detection performance, efficiency, and cross-domain generalizability of \ourmethod.



%% file: sections/1_intro.tex
Tabular data anomaly detection (TAD) aims to identify anomalous samples that are significantly different from the majority of samples in tabular data~\cite{borisov2022deep,shenkar2022anomaly,thimonier2023beyond}. TAD has broad applications across various real-world scenarios, such as rare disease diagnosis~\cite{fernando2021deep}, network intrusion detection~\cite{ahmad2021network} and financial fraud detection~\cite{al2021financial}. To date, most existing TAD methods have demonstrated strong performance under the \textbf{dataset-specific paradigm}, where anomaly detection model training and inference are conducted on the same dataset. Despite their effectiveness, this paradigm requires training a separate detector for each new dataset and usually relies on dataset-specific hyperparameter tuning, leading to high computational and deployment costs~\cite{ruff2021unifying,golan2018deep}. Moreover, these dataset-specific detectors tend to overfit to the training distribution and struggle to transfer to unseen datasets, which hinders the scalability and general applicability of TAD systems.~\cite{yin2024mcm,ye2025drl,goodge2022lunar}.


\begin{figure} [t]
    \centering
    \subfigure[Feature alignment comparison]{\label{subfig:intro_1}
        \includegraphics[height=2.8cm]{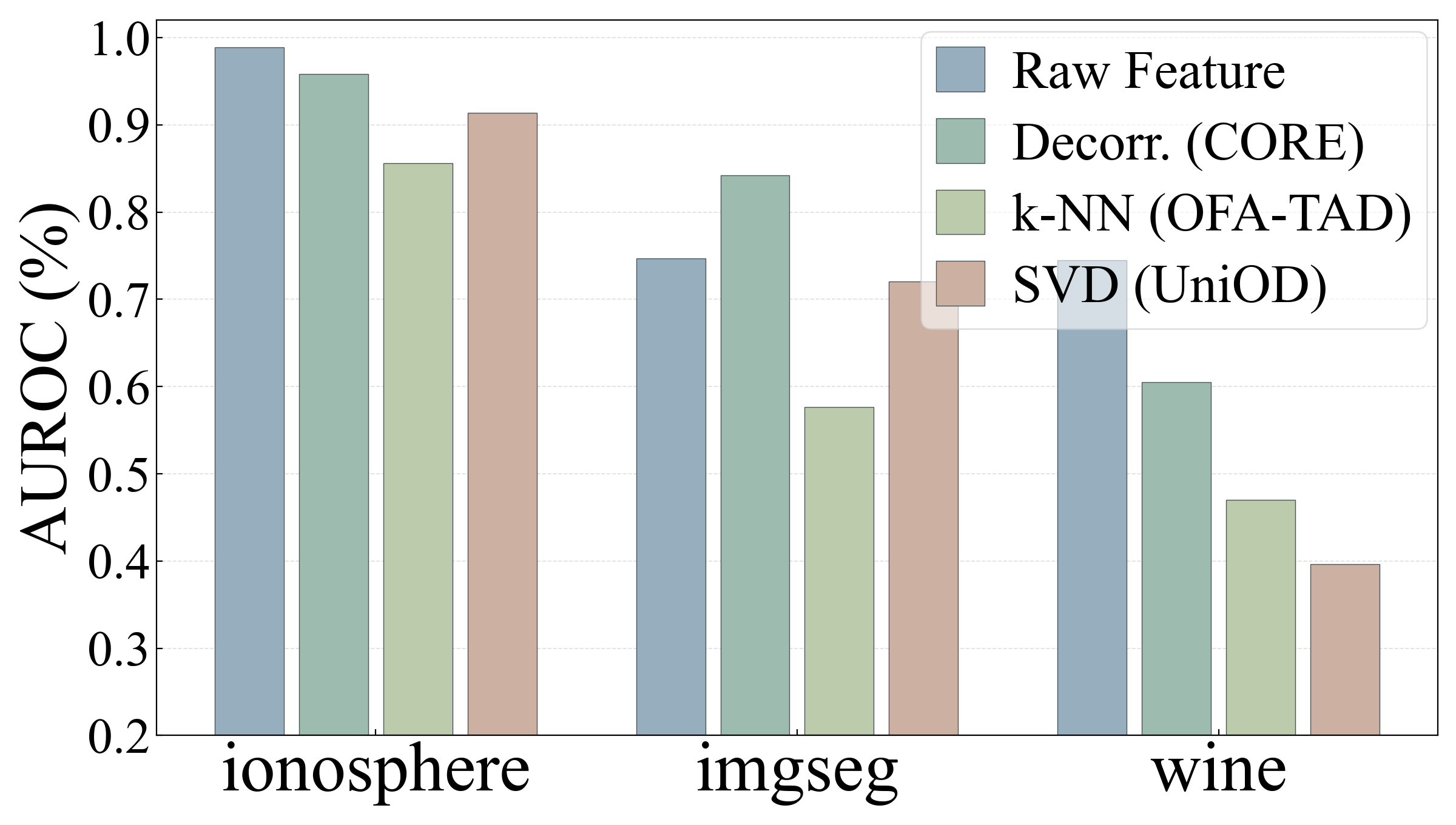}
    }\hfill
    \subfigure[Synthetic anomalies]{\label{subfig:intro_2}
        \includegraphics[height=2.8cm]{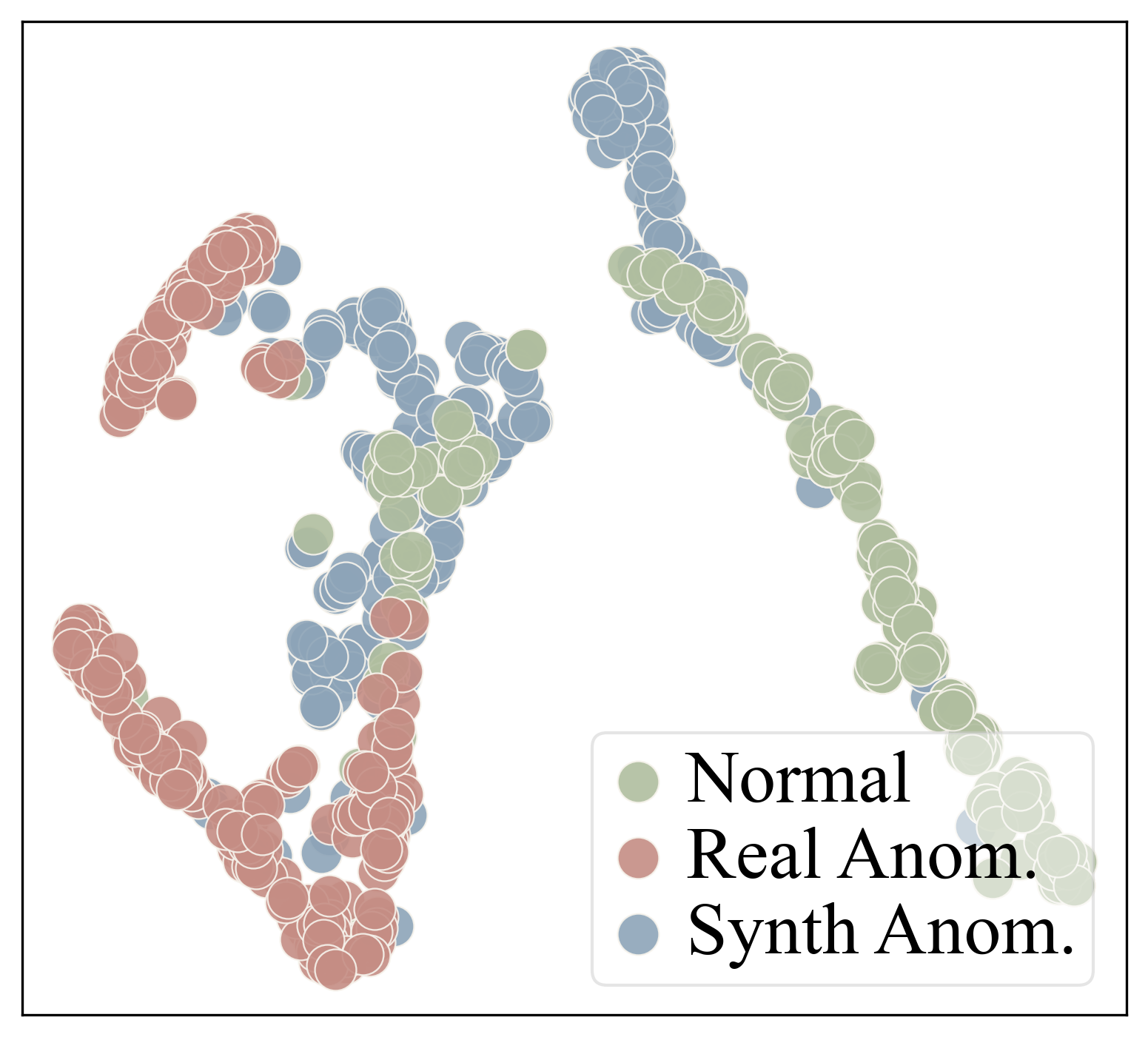}
    }
    \vspace{-2mm}
    \caption{(a): Performance comparison of different feature alignment strategies. (b): T-SNE visualization of normal samples, synthetic anomalies, and real anomalies learned by OFA-TAD on breastw dataset.}
    \label{fig:intro}
    \vspace{-1mm}
\end{figure}

 

Recently, tabular foundation models, which aim to develop a unified model for diverse tabular datasets, have achieved remarkable progress in predictive tasks, such as classification, regression, and missing value imputation~\cite{hollmann2025accurate,ma2026tabdpt}. 
Following this trend, a promising direction for TAD is to embrace the \textbf{foundation model paradigm}, where a single unified detector can generalize to identify anomalies for various tabular datasets from different application domains. In this context, several pioneering studies, such as UniOD~\cite{fu2025uniod} and OFA-TAD~\cite{li2026towards}, have explored \textbf{unified TAD} approaches, addressing the above shortages by reducing deployment costs and improving generalizability to different applications. 


Despite its merits, the development of unified TAD models remains non-trivial due to several unique challenges that do not arise under the dataset-specific paradigm. A primary challenge is \textit{\textbf{Challenge 1 - unification of heterogeneous data}}. Due to differences in data characteristics across domains, the dimensionality and feature semantics of tabular datasets can vary significantly. 
To support cross-domain transferability in unified models, heterogeneous tabular inputs need to be mapped into a shared representation space. To achieve this goal, UniOD~\cite{fu2025uniod} derives unified representations by factorizing multi-scale pairwise similarity matrices, while OFA-TAD~\cite{li2026towards} represents each sample using its $k$-NN distance to other samples as unified input. Although these alignment strategies are anomaly-oriented, their aligned features are indirectly constructed rather than directly derived from the original features, potentially causing the \textit{loss of original feature semantics}. For example, in financial transaction datasets, losing the semantics of transaction amount and merchant type can make suspicious and normal purchases harder to distinguish, leading to sub-optimal detection performance. As shown in Fig.~\ref{subfig:intro_1}, under the same reconstruction-based detector, indirect alignment strategies (i.e., $k$-NN and SVD) consistently achieve worse performance compared with using the raw features, indicating their information loss.

In addition to data alignment, another key challenge for unified TAD is \textit{\textbf{Challenge 2 - unification of anomaly detection objectives}}. Unlike other predictive tasks, anomaly detection is difficult to unify, as anomalous patterns in real-world data can differ substantially across application domains. To address this challenge, existing approaches typically formulate unified TAD as a binary classification task~\cite{fu2025uniod,li2026towards}, which may overlook the diverse nature of anomalies. On the one hand, treating anomalies as a single class is potentially inadequate, as they exhibit diverse patterns and often fail to form a semantically consistent cluster~\cite{liznerski2020explainable}. On the other hand, to train a unified detector with scarce real-world data, some unified TAD methods rely on data synthesis to generate pseudo-anomalies~\cite{li2026towards} or even entire training datasets~\cite{ding2026zero}. In this case, the synthetic data may fail to reflect the true distribution of real-world scenarios, which limits their detection performance in practical applications. As visualized in Fig.~\ref{subfig:intro_2}, the representations of synthetic anomalies in OFA-TAD~\cite{li2026towards} significantly deviate from those of real anomalies, making the learned classification boundary less reliable. 


By rethinking the solutions for aforementioned challenges, in this paper, we propose a in-\textbf{CO}ntext \textbf{RE}construction approach for unified TAD, abbreviated as \textbf{\ourmethod}. 
To address \textit{\textbf{Challenge 1}}, we introduce a \textit{decorrelated feature alignment module}, which directly transforms heterogeneous features into a unified representation space. Specifically, \ourmethod first selects informative and less redundant features as representative attributes and then arranges them in a consistent order for alignment. In this way, the aligned features achieve dimensional consistency while preserving the semantics of the original features. 
To handle \textit{\textbf{Challenge 2}}, we depart from classification objective and reformulate unified TAD as a reconstruction problem. To adapt reconstruction model to unified TAD paradigm, we propose a \textit{in-context reconstruction module} that reconstructs each sample using auxiliary normal samples as references in an in-context learning manner. Concretely, \ourmethod adaptively retrieves the most relevant normal samples to guide the reconstruction of each query; as a result, the reconstruction model can adapt to any unseen distributions, yielding reconstruction errors to reliably reflect abnormality. 
This paper makes the following contributions:

\begin{itemize}
    \item \textbf{New Formulation.} To our knowledge, we are the first to formulate unified TAD as a reconstruction problem, aiming to detect anomalies in any dataset without relying on labeled or synthesized anomalies.

    \item \textbf{Novel Method.} We propose \ourmethod, which directly aligns heterogeneous features via decorrelated feature alignment and detects anomalies via in-context reconstruction.

    \item \textbf{Extensive Experiments.} We conduct experiments on 34 datasets from diverse domains, demonstrating the effectiveness, efficiency, and generalizability of \ourmethod.
\end{itemize}

%% file: sections/2_rw.tex

\noindent\textbf{Tabular Data Anomaly Detection (TAD)} aims to detect anomalous samples that deviate significantly from the majority normal patterns in tabular data~\cite{parzen1962estimation,kim2023odim,pimentel2014review}. Earlier studies primarily focused on classical machine learning approaches for TAD~\cite{scholkopf2001estimating,li2022ecod}. For example, Isolation Forest~\cite{liu2008isolation} isolates anomalies through random partitioning, while Local Outlier Factor~\cite{breunig2000lof} identifies anomalies by comparing the local density of each sample with that of its neighbors. Recent research highlights the effectiveness of deep learning in TAD~\cite{pang2021deep,kim2019rapp,livernoche2024diffusion}. For instance, AE~\cite{chen2018autoencoder} trains an encoder-decoder network to reconstruct normal samples and uses the reconstruction error as the anomaly score, while MCM~\cite{yin2024mcm} extends this idea with masked cell modeling to capture feature correlations through masked reconstruction. Apart from reconstruction-based models, NeuTraLAD~\cite{qiu2021neural} uses learnable data transformations to maps features into a representation space with similar embeddings. However, these methods still follow the data-specific paradigm for training and inference, making them struggle to generalize into unseen datasets due to distribution shifts between datasets.
%

\noindent\textbf{Unified TAD} methods aim to identify anomalies across diverse tabular datasets with a single generalist model, without dataset-specific training or fine-tuning~\cite{fu2025uniod}. Recent studies have explored in-context prediction for this task. For instance, FoMo-0D~\cite{shen2025fomo} formulates tabular outlier detection as a zero-shot prediction problem and directly identifies anomalies in unseen datasets using context samples. OUTFORMER~\cite{ding2026zero} further advances FoMo-0D by enriching synthetic detection tasks and adaptively organizing them during pre-training. Considering the inconsistent feature dimensions and semantics across tabular datasets, another line of studies constructs dataset-agnostic representations. UniOD~\cite{fu2025uniod} constructs kernel-based similarity graphs with SVD-derived node embeddings for cross-domain anomaly detection. OFA-TAD~\cite{li2026towards} extracts multi-view neighbor-distance profiles and fuses view-specific anomaly evidence with a Mixture-of-Experts network. 
 Despite their effectiveness, existing unified TAD methods mainly formulate anomaly detection as a binary classification task, which may overlook the diverse nature of anomalies across application domains and be misled by synthetic data. 
 

A more extensive literature review is in Appendix A.

%% file: sections/3_pre.tex
\noindent \textbf{Notations.} A tabular dataset can be denoted as $\mathcal{D} = (\mathbf{X}, \mathbf{y})$, where $\mathbf{X} \in \mathbb{R}^{n \times d}$ represents the feature matrix with $n$ samples and $d$ features, and its $i$-th row $\mathbf{x}_i\in\mathbb{R}^{d}$ denotes the feature vector of the $i$-th sample. In the context of anomaly detection, the labels are denoted by $\mathbf{y} \in \{0, 1\}^n$, where $y_i=1$ and $y_i=0$ represent anomalous and normal samples, respectively. Following the one-class TAD setting, the training set $\mathcal{D}_{\mathrm{train}}$ contains only normal samples, while the test set $\mathcal{D}_{\mathrm{test}}$ consists of both normal and anomalous samples. The objective of TAD is to learn an anomaly scoring function $f: \mathbb{R}^d \rightarrow \mathbb{R}$, where higher scores indicate higher abnormality, i.e., $f(\mathbf{x}_a)>f(\mathbf{x}_n)$ for anomalous sample $\mathbf{x}_a$ and normal sample  $\mathbf{x}_n$.

\noindent \textbf{Unified TAD Problem.} Following~\cite{li2026towards}, we consider the unified TAD setting where a model is trained on source datasets and directly applied to unseen target datasets without retraining. Let $\mathbb{D}_{\mathrm{source}}$ and $\mathbb{D}_{\mathrm{target}}$ denote the source and target dataset collections with $\mathbb{D}_{\mathrm{source}} \cap \mathbb{D}_{\mathrm{target}} = \emptyset$. For each dataset $\mathcal{D}_j$, we denote its training and test splits as $\mathcal{D}_{j,\mathrm{train}}$ and $\mathcal{D}_{j,\mathrm{test}}$. The training splits of all source datasets are collected to construct the source training pool, i.e., $\mathcal{P}_{\mathrm{train}}=\bigcup_{\mathcal{D}_j\in\mathbb{D}_{\mathrm{source}}}\mathcal{D}_{j,\mathrm{train}}$. The unified detector $f(\cdot)$ is optimized on $\mathcal{P}_{\mathrm{train}}$. During evaluation, for each target dataset $\mathcal{D}_k\in\mathbb{D}_{\mathrm{target}}$, $f(\cdot)$ predicts anomaly scores on $\mathcal{D}_{k,\mathrm{test}}$ (i.e., query samples) by using $\mathcal{D}_{k,\mathrm{train}}$ (i.e., context samples) as the target-domain context.


%% file: sections/4_method.tex
\begin{figure*}
    \centering
        \includegraphics[width=.98\linewidth]{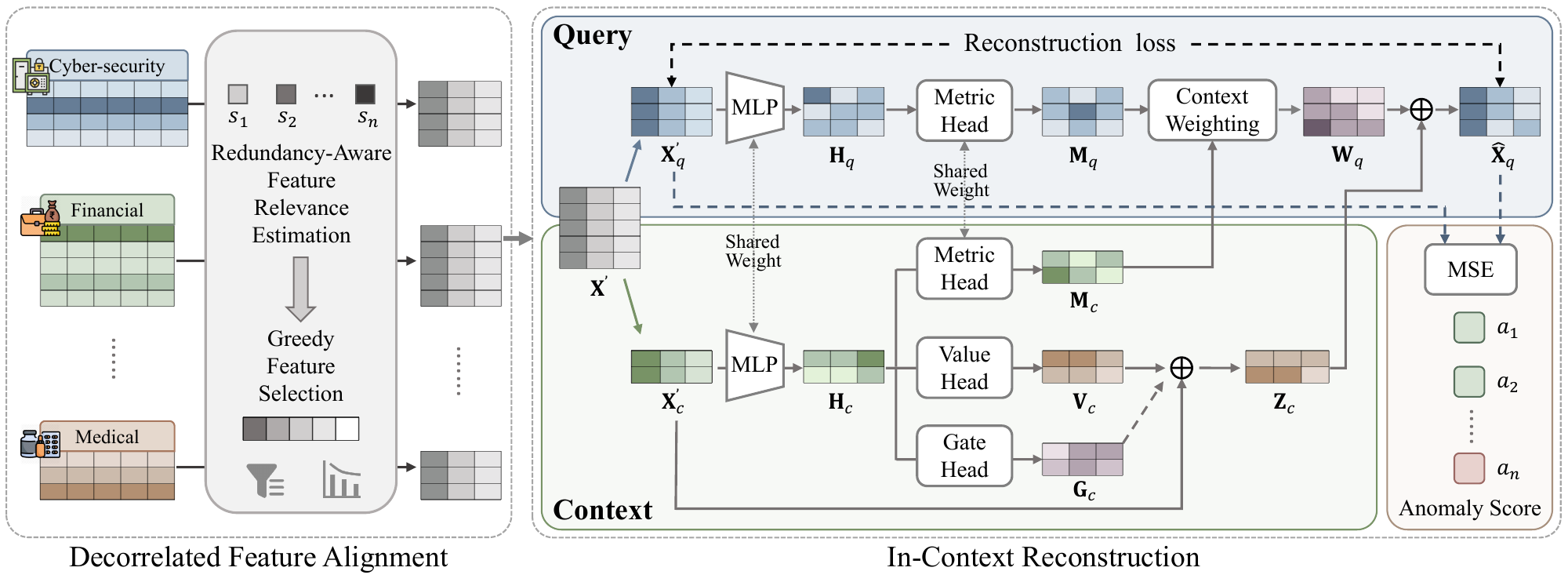}
    \caption{The overall pipeline of \ourmethod. The \textit{decorrelated feature alignment} module greedily selects informative and less redundant original features and arranges them into a unified representation. Then, the \textit{in-context reconstruction} module reconstructs each query according to its relevant normal context samples. The reconstruction errors serves as anomaly scores.}
	\label{fig:pipeline}
\end{figure*}

In this section, we introduce \ourmethod, a in-\textbf{CO}ntext \textbf{RE}construction approach for unified TAD. The overall pipeline of \ourmethod is illustrated in Fig.~\ref{fig:pipeline}. First, to align heterogeneous tabular data from diverse domains, we introduce a \textit{Decorrelated Feature Alignment} module, which greedily selects features with high variance and low inter-feature correlation, and arranges them into a unified representation. Next, to estimate query abnormality with context samples, we propose a \textit{In-Context Reconstruction} module, which reconstructs each query from its most relevant context samples through a gated aggregation mechanism. Ultimately, the reconstruction error between each query and its reconstructed representation serves as the anomaly score.

\subsection{Decorrelated Feature Alignment}\label{subsec:align}
To establish a unified TAD model, the first challenge is to unify heterogeneous tabular data, where feature dimensionality and semantics differ significantly across domains and datasets. To make them compatible with the input of a unified model, in the first step, we need to align heterogeneous features into a unified representation space. Facing this challenge, most existing unified TAD methods~\cite{li2026towards,fu2025uniod} align heterogeneous features through indirect feature transformations, such as neighbor-distance encoding and graph-based decomposition, which introduces two limitations. Firstly, although indirect feature transformations can align dimensionality, they fail to preserve the semantic meaning of features across datasets. 
Secondly, they ignore feature redundancy when constructing unified representations. In tabular data, strongly correlated attributes (e.g., annual income and monthly income) may provide overlapping information, while less correlated attributes (e.g., age and income) can capture more distinct aspects for anomaly detection. Together, these limitations can lead to unified representations that are both semantically misaligned and information-redundant, resulting in limited representational capacity for anomaly detection. 


To address the above limitations, we introduce a decorrelated feature alignment module in \ourmethod, which directly constructs unified representations from the original features while reducing feature redundancy. Specifically, the module first measures the contribution of each feature through redundancy-aware relevance scoring, and then greedily selects and ranks features according to their informativeness to construct the aligned representation.

\noindent\textbf{Redundancy-Aware Feature Relevance Estimation.}
The key idea of the alignment module in \ourmethod is to construct a fixed-dimensional representation by selecting original features according to their informativeness and redundancy. To quantify these properties, 
we introduce a relevance score that jointly considers feature variance and inter-feature correlation. Specifically, given a feature matrix $\mathbf{X} \in \mathbb{R}^{n \times d}$, we use the variance $\operatorname{Var}(\mathbf{X}_j)$ to measure the informativeness of feature $j$, and estimate its redundancy by the maximum absolute correlation with the remaining features:

\begin{equation} 
\rho_j^{\operatorname{max}} = \operatorname{max}_{k \neq j} |\operatorname{Corr}(\mathbf{X}_j, \mathbf{X}_k)|, 
\end{equation}

\noindent where $\operatorname{Corr}(\cdot)$ denotes the Pearson correlation coefficient, and $\rho_j^{\operatorname{max}}$ denotes the maximum redundancy of feature $j$ with respect to the remaining features. Accordingly, the relevance score of feature $j$ is defined as:

\begin{equation} 
s_j = \operatorname{Var}(\mathbf{X}_j) \cdot (1 - \rho_j^{\operatorname{max}}), 
\end{equation}  
\noindent where $s_j$ denotes the relevance score of feature $j$, with a higher $s_j$ indicating greater informativeness and lower redundancy with respect to the other features. 

\noindent\textbf{Greedy Feature Selection.} 
Using the above relevance score, a straightforward strategy is to rank features according to $s_j$ in a static way and select the top-ranked ones for as aligned feature. 
However, even if a group of features is highly correlated, one of them may still provide useful discriminative information for anomaly detection. In this case, pure relevance-based selection may assign similarly low scores to all features in the group, causing them to be ranked below less informative but weakly correlated features. As a result, the selected subset may discard representative features that are important for anomaly detection.


To address this issue, we introduce a greedy feature selection strategy that updates feature redundancy with respect to the already selected feature set, ensuring that only features redundant to the selected representatives are penalized. 
Specifically, we maintain a selected set $\mathcal{S}$, initialized as empty, containing all selected features. At each iteration, we compute the score of each candidate feature $j$ by estimating its relevance with the selected features:

\begin{equation}
s_j^{*} = \operatorname{Var}(\mathbf{X}_j) \cdot \left(1 - \operatorname{max}_{k \in \mathcal{S}} |\operatorname{Corr}(\mathbf{X}_j, \mathbf{X}_k)|\right),
\end{equation}

\noindent where $\mathcal{S}$ denotes the set of selected features. Then, the feature with the highest score is selected and added to $\mathcal{S}$:

\begin{equation}
j^{\prime} = \operatorname{argmax}_{j \notin \mathcal{S}}\, s_j^{*}, \quad \mathcal{S} \leftarrow \mathcal{S} \cup \{j^{\prime}\}.
\end{equation}

\noindent where $j^{\prime}$ denotes the index of the highest-scoring feature selected at each iteration. The above selection step is repeated until $|\mathcal{S}|$ reaches the target dimension $d_u$. Consequently, one representative feature from each correlated group is retained, rather than suppressing all correlated features simultaneously. 

After feature selection, we arrange them according to their selected order. For each dataset, features selected earlier are placed in the front dimensions, while features selected later are placed at the end. In this way, each dataset is transformed into $\mathbf{X}' \in \mathbb{R}^{n \times d_u}$ with a unified dimensionality and a consistent feature order. In practice, when the original dimensionality $d$ of a dataset is smaller than $d_u$, we apply a random projection to map the original features into a higher-dimensional space before performing feature selection.

\subsection{In-Context Reconstruction}\label{subsec:recon}
With the unified feature, the remaining challenge is to unify the anomaly detection for different datasets into a unified task. While classification-based formulation may oversimplify the diverse nature of anomalies, a promising alternative is to adopt a reconstruction-based formulation, which has been widely used for anomaly detection for various data modalities~\cite{ding2019deep,zavrtanik2021reconstruction}. 
In reconstruction-based anomaly detection, the reconstruction model (e.g. autoencoders) captures normal patterns by learning to reconstruct normal samples; then, the reconstruction errors (i.e., MSE) can serve as anomaly scores, as anomalous samples typically deviate from the learned normality and show higher errors. 




Follow the dataset-specific paradigm, conventional reconstruction-based methods only need to learn the normal patterns of a specific dataset. 
However, in the unified TAD setting, normal patterns may vary across datasets, making it difficult for a unified reconstruction model to capture transferable patterns from different domains. For example, normal samples from medical and financial datasets may exhibit fundamentally different feature distributions and dependencies, such that patterns learned from one domain may not characterize normality in another. 
To address this issue, we introduce a \textit{in-context reconstruction module} that learns to reconstruct based on the normal patterns inferred from dataset-specific contexts. 
Concretely, we treat a subset of normal samples as \textbf{context samples}, that provide dataset-specific normal references, to guide the reconstruction of unlabeled \textbf{query samples}. 
For each query sample, we retrieve its most relevant context samples, and aggregate them through a gated aggregation mechanism for query reconstruction. The detailed reconstruction process is described as follows.


\noindent\textbf{Multi-Head Encoding.}
To explore contextual information during the reconstruction process, we need to {\textit{retrieve appropriate context samples} for a specific query}, {\textit{extract informative representations} for reconstruction}, and {\textit{preserve query-specific information} from their original features}. 
To fulfill these distinct functions, we devise a multi-head encoding block that maps context samples into role-specific representations. 
Specifically, given the aligned feature matrix $\mathbf{X}'$, we partition it into two parts: the context set $\mathbf{X}'_c \in \mathbb{R}^{n_c \times d_u}$ and the query set $\mathbf{X}'_q \in \mathbb{R}^{n_q \times d_u}$. Then, both sets are fed into a shared backbone to obtain their hidden representations:

\begin{equation}
\mathbf{H}_q = f_\theta(\mathbf{X}'_q), \quad
\mathbf{H}_c = f_\theta(\mathbf{X}'_c),
\end{equation}

\noindent where $f_\theta(\cdot)$ is a two-layer MLP. After obtaining these representations, we apply three lightweight heads $g(\cdot)$ to extract the role-specific information from each context sample:

\begin{equation}
\mathbf{M}_c = g_{\mathrm{m}}(\mathbf{H}_c), \quad
\mathbf{V}_c = g_{\mathrm{v}}(\mathbf{H}_c), \quad
\mathbf{G}_c = g_{\mathrm{g}}(\mathbf{H}_c),
\end{equation}

\noindent where $\mathbf{M}_c$, $\mathbf{V}_c$, and $\mathbf{G}_c$ denote the metric representations for measuring query-context similarity, the value representations for reconstruction, and the dimension-wise gate representations for feature fusion, respectively. Meanwhile, query samples are encoded by the metric head as $\mathbf{M}_q = g_{\mathrm{m}}(\mathbf{H}_q)$ to retrieve the most relevant context samples.

\noindent\textbf{Context Weighting.}
After obtaining the role-specific representations of query and context samples, we reconstruct each query from the context samples. Nevertheless, given a specific query, not all context samples are equally relevant to it for reconstruction. Directly aggregating all context samples may introduce dissimilar patterns into the reconstruction, degrading the quality of the reconstructed query. To address this issue, we retrieve the most relevant context samples for each query before reconstruction. Concretely, for the $i$-th query sample, we compute the Euclidean distance $d_{i,k} = \|\mathbf{m}_{q,i}-\mathbf{m}_{c,k}\|_2$ between its metric representation $\mathbf{m}_{q,i}$ and the corresponding context representation $\mathbf{m}_{c,k}$. Then, we select the $K$ nearest context samples for the $i$-th query and conduct a Softmax-based normalization to obtain their contribution weights:

\begin{equation}
w_{i,k} = \frac{\exp\!\big(-d_{i,k} / \tau\big)} {\sum_{j=1}^{K} \exp\!\big(-d_{i,j} / \tau\big)},
\end{equation}

\noindent where $k \in \{1, \dots, K\}$ indexes the selected context samples, $w_{i,k}$ denotes the contribution weight of the $k$-th selected context sample for the $i$-th query, and $\tau$ is the temperature hyperparameter to adjust the smoothness of the weight distribution. 
The weighting mechanism supports adaptive context aggregation, allowing each query to emphasize the most informative context samples during reconstruction. 

\noindent\textbf{Gated Fusion.}
To reconstruct each query based on the selected context samples, a straightforward strategy is to directly use their value representations for reconstruction. 
However, the value representations are obtained after multi-layer transformations, which may introduce noise and blur anomaly-related patterns. 
To address this issue, we employ a dimension-wise gate to combine the original aligned feature and the value representation of each selected context sample. 
Specifically, for query $\mathbf{x}_i$, the fused representation its $k$-th selected context sample is computed as:

\begin{equation}\label{eq_gate}
\mathbf{z}_{i,k} = {\sigma}(\mathbf{g}_{i,k}) \odot \mathbf{x}'_{i,k}
+ \big(1 - {\sigma}(\mathbf{g}_{i,k})\big) \odot \mathbf{v}_{i,k},
\end{equation}

\noindent{where $\mathbf{x}'_{i,k}$, $\mathbf{v}_{i,k}$, and $\mathbf{g}_{i,k}$ denote the aligned feature, value representation, and gate vector of the $k$-th selected context sample for query $i$, respectively, retrieved from the corresponding rows of $\mathbf{X}'_c$, $\mathbf{V}_c$, and $\mathbf{G}_c$. 
Their first subscript indexes the query sample, while the second indexes the retrieved context sample for that query. 
${\sigma}(\cdot)$ is the sigmoid function, and $\odot$ denotes element-wise multiplication. 
The gating mechanism ensures that the fused representation can adaptively balance the aligned feature and value representation in each dimension. When the original features are more useful for a certain feature dimension, the fused representation retains more information from the aligned feature; otherwise, it relies more on the value representation for reconstruction. 
In this way, each selected context sample can provide a more suitable representation for reconstructing the query.

\noindent\textbf{Reconstruction Scoring.}
After obtaining the fused representations of the selected context samples, we reconstruct the query through their weighted combination:

\begin{equation}
\hat{\mathbf{x}}_i = \sum_{k=1}^{K} w_{i,k}\, \mathbf{z}_{i,k},
\end{equation}

\noindent where $\hat{\mathbf{x}}_i$ denotes the reconstructed features. Then, the anomaly score $a_i$ for query sample $\mathbf{x}_i$ is calculated by the mean squared reconstruction error:

\begin{equation}
a_i = \frac{1}{d}\big\| \mathbf{x}'_i - \hat{\mathbf{x}}_i \big\|_2^2.
\end{equation}

%

This score measures the reconstruction deviation between the aligned feature and its reconstruction, with a larger value indicating that the query sample is more likely to be abnormal. To optimize \ourmethod, we minimize the reconstruction error on the training sets of source datasets. The algorithm and complexity analysis of \ourmethod are provided in Appendix B.

%% file: sections/5_exp.tex
\input{tables/main_auroc}

\subsection{Experimental Setup}
\noindent\textbf{Datasets.} 
Following~\cite{li2026towards}, we train \ourmethod on 7 source datasets and test on 34 target datasets. The source datasets are selected from diverse domains, including Astronautics (satellite), Biology (vertebral), Healthcare (annthyroid and Cardiotocography), Image (imgseg), Physical \& Chemistry (wine), and Sociology (comm.and.crime), while the remaining datasets are used as target datasets. Among the target datasets, those sharing the same categories with the source datasets are treated as in-domain datasets, while the others are treated as out-of-domain datasets. Detailed dataset statistics and splits are reported in Appendix C.1.

\noindent\textbf{Baselines.}
We compare \ourmethod with 10 representative TAD methods, including classic machine learning methods, i.e., Isolation Forest (IForest)~\cite{liu2008isolation}, Local Outlier Factor (LOF)~\cite{breunig2000lof}, and kNN-based detection (KNN)~\cite{angiulli2002fast}; deep learning-based approaches, i.e., AutoEncoder (AE)~\cite{chen2018autoencoder}, DeepSVDD (DSVDD)~\cite{liznerski2020explainable}, LUNAR~\cite{goodge2022lunar}, MCM~\cite{yin2024mcm}, DRL~\cite{ye2025drl}, and DisentAD~\cite{ye2025disentangling}; and the unified TAD method, i.e., OFA-TAD~\cite{li2026towards}. For detailed information, refer to Appendix C.2.

\noindent\textbf{Evaluation Metrics and Implementation.} 
Following~\cite{li2026towards}, we employ AUROC and AUPRC as primary metrics, with F1 score reported as an additional threshold-dependent metric, and repeat each experiment over 5 trials. During training, for each source dataset, we randomly sample 15\% of its training samples as context samples and use the remaining training samples as query samples. During inference, all training samples of each target dataset are used as context samples, with test set as queries. The hyperparameters are kept fixed across all target datasets without dataset-specific tuning. For baselines, we follow the results and evaluation protocol reported in OFA-TAD~\cite{li2026towards}. More implementation details are given in Appendix C.3.

\subsection{Experimental Results}
\noindent\textbf{Performance Comparison.} 
Table~\ref{tab:main_results} shows the AUROC comparison between \ourmethod and baseline methods on 34 target datasets, and AUPRC/F1 results are deferred to Appendix D.1. We have the following observations: \ding{182} \ourmethod demonstrates strong anomaly detection capability in the generalist TAD scenario. Specifically, \ourmethod achieves the best average AUROC of 0.8488, outperforming the second-best method OFA-TAD by 1.43 percentage points. Moreover, \ourmethod performs better than OFA-TAD on 26 out of 34 target datasets, indicating that the improvement is consistent rather than driven by only a few datasets; \ding{183} Even on the out-of-domain target datasets, \ourmethod remains robust under domain shift. On the 11 out-of-domain datasets, \ourmethod achieves an average AUROC of 0.8178, which improves over the second-best method OFA-TAD by 2.59 percentage points. The consistent advantage under unseen domains highlights the effectiveness of our feature alignment strategy and context-query reconstruction mechanism for cross-domain TAD; \ding{184} \ourmethod also achieves the best average rank among all compared methods, indicating that its advantage is not only reflected in average AUROC but also in cross-dataset stability. 
These results show that \ourmethod provides a more balanced trade-off between overall detection performance and cross-dataset robustness.

\input{tables/ablation}
\noindent\textbf{Ablation Study.} 
To verify the effectiveness of each key component of \ourmethod, we design three variants: 
\ding{182}~\textbf{w/o Align}, using random projection to replace decorrelated feature alignment; 
\ding{183}~\textbf{w/o Gating}, removing the adaptive gating for reconstruction; and 
\ding{184}~\textbf{w/o Context}, replacing in-context reconstruction with autoencoder-style self-reconstruction. 
The results are shown in Table~\ref{tab:ablation}. We observe that all three components contribute to the performance of \ourmethod. Among them, removing in-context reconstruction (\textbf{w/o Context}) causes the largest drop, suggesting that reconstructing queries from the most relevant context samples is crucial for adapting to diverse normal patterns across datasets. Removing the gating module (\textbf{w/o Gating}) also leads to a clear degradation, indicating that adaptive fusion of aligned features and value representations helps build context representations for reconstruction. In addition, replacing decorrelated feature alignment with random projection (\textbf{w/o Align}) causes a mild but consistent degradation, suggesting that selecting informative yet less redundant features helps construct a stable unified representation under cross-dataset feature heterogeneity.

\input{figs/ContextRatio/contextRatio}
\noindent\textbf{Sensitivity of \#Context Samples.} 
Fig.~\ref{fig:context_ratio} shows the impact of varying context ratio on the average anomaly detection performance of \ourmethod and OFA-TAD. Note that the context ratio ranges from 0.1 to 1.0 on target training set. It can be observed that both methods benefit from a larger context ratio, indicating that more context samples provide richer normal patterns for reconstruction. Furthermore, we can conclude that \ourmethod consistently outperforms OFA-TAD under different context ratios in terms of both AUROC and AUPRC, which further demonstrates the effectiveness of the proposed in-context reconstruction framework. Additional context ratio results on specific datasets are provided in Appendix D.2.

\input{figs/Efficiency/eff}
\noindent\textbf{Efficiency Analysis.} 
To assess the efficiency of \ourmethod, we compare the whole runtime of \ourmethod with baseline methods on the Wilt dataset. From the results in Fig.~\ref{fig:efficiency} (more results are in Appendix D.3), we can see \ourmethod achieves the lowest runtime among all compared methods. Specifically, \ourmethod demonstrates comparable or better runtime performance than the fastest classical methods, such as LOF and KNN, and significantly outperforms most deep learning-based methods. Different from most baseline methods, \ourmethod directly performs inference on the target dataset without target-specific training or fine-tuning, and thus avoids heavy running cost. 

\input{figs/ScalingLaw/scale}
\noindent\textbf{Scaling with Source Data and Model Size.} 
We further investigate the scaling behavior of \ourmethod by varying the amount of source training data and the model size. Specifically, we scale the number of source training samples from 0.13K to the full training set, and increase the number of model parameters from 0.02M to 3.00M. As shown in Fig.~\ref{fig:scale}, \ourmethod shows a clear scaling trend: increasing source data consistently improves the generalization ability, and larger models generally achieve better results when sufficient data are available. 
This suggests that more source samples provide richer reconstruction cases across heterogeneous tabular datasets, while larger models offer stronger capacity to capture transferable patterns for in-context reconstruction. 
These results demonstrate the scalability of \ourmethod and its potential for further improvement with larger-scale pre-training.

%% file: tables/main_auroc.tex
\begin{table*}[t]
\centering
\small
\setlength{\tabcolsep}{6.0pt}
\begin{tabular}{l|ccc|cccccc|c|>{\columncolor{corecolbg}}c}
\toprule
Dataset & LOF & KNN & iForest & DSVDD & AE & MCM & LUNAR & DRL & DisentAD & OFA-TAD & \ourmethod \\
\midrule
\rowcolor{gray!15}
\multicolumn{12}{c}{\textbf{In-Domain Target Datasets}} \\
\midrule
abalone      & 72.84 & 79.97 & 73.71 & 67.56 & 80.14 & 74.50 & \secondplace{80.84} & \thirdplace{80.71} & 77.89 & \firstplace{81.78} & 79.07 \\
arrhythmia   & \thirdplace{80.75} & 80.14 & 79.31 & 75.66 & 72.60 & 78.48 & 76.23 & 74.67 & 75.67 & \secondplace{80.95} & \corefirst{81.88} \\
breastw      & 97.76 & 97.64 & \secondplace{99.75} & 99.13 & 97.78 & \firstplace{99.77} & 98.36 & 99.49 & \thirdplace{99.60} & 97.91 & 99.39 \\
cardio       & 93.74 & 90.62 & 93.52 & \firstplace{95.71} & \secondplace{95.00} & 87.10 & 91.21 & 94.31 & \thirdplace{94.80} & 93.22 & 94.63 \\
census       & 54.51 & 67.51 & 63.64 & \thirdplace{69.16} & \thirdplace{69.16} & 66.96 & 65.65 & 59.14 & \firstplace{83.45} & \secondplace{69.55} & 69.08 \\
donors       & 98.45 & \thirdplace{99.91} & 90.29 & 74.94 & 93.84 & 99.65 & 99.79 & 90.02 & 90.73 & \secondplace{99.97} & \corefirst{100.00} \\
fault        & 47.42 & 58.73 & 56.65 & 51.63 & 58.18 & \firstplace{61.65} & 53.40 & \thirdplace{59.10} & \secondplace{61.44} & 57.62 & 58.86 \\
Hepatitis    & 61.31 & 49.10 & 72.26 & \firstplace{78.37} & 56.56 & 61.08 & 55.34 & 68.55 & 63.71 & \thirdplace{73.53} & \coresecond{76.83} \\
lympho       & 95.77 & 93.43 & \thirdplace{99.72} & \secondplace{99.81} & 94.13 & 99.15 & 98.28 & 99.20 & 83.38 & 99.11 & \corefirst{100.00} \\
mammography  & 85.99 & 87.24 & 88.35 & 86.15 & 89.35 & \secondplace{90.73} & 87.19 & 87.88 & 88.80 & \thirdplace{90.00} & \corefirst{90.81} \\
mnist        & \thirdplace{95.11} & 93.48 & 86.57 & 84.34 & 92.15 & \secondplace{96.40} & 87.96 & \firstplace{96.45} & 58.35 & 94.33 & 95.02 \\
musk         & \firstplace{100.00} & \firstplace{100.00} & 97.48 & 98.25 & \firstplace{100.00} & 99.66 & \thirdplace{99.75} & \secondplace{99.99} & 96.90 & \firstplace{100.00} & \corefirst{100.00} \\
optdigits    & \firstplace{99.36} & 96.80 & 80.20 & 56.36 & 85.35 & 98.91 & 93.29 & 82.25 & \thirdplace{99.26} & \secondplace{99.30} & 99.07 \\
Parkinson    & 69.67 & 46.15 & \firstplace{77.18} & 68.40 & 69.36 & 33.79 & 47.46 & 66.03 & 71.01 & \thirdplace{71.64} & \coresecond{73.86} \\
pendigits    & 98.71 & 98.83 & 96.42 & 77.54 & 95.54 & 98.42 & 98.97 & 93.91 & \thirdplace{99.32} & \firstplace{99.90} & \coresecond{99.89} \\
pima         & 66.09 & 68.07 & \secondplace{73.31} & 68.98 & 72.82 & 71.31 & \firstplace{73.49} & \thirdplace{73.13} & 71.61 & 69.59 & 72.50 \\
satimage-2   & 99.38 & \thirdplace{99.71} & 99.10 & 97.30 & \secondplace{99.85} & \firstplace{99.86} & 95.17 & 98.57 & 85.15 & 99.64 & 99.66 \\
shuttle      & 99.72 & 99.64 & 99.64 & 99.34 & 99.76 & 99.86 & 97.59 & 99.83 & \thirdplace{99.93} & \secondplace{99.98} & \corefirst{99.99} \\
thyroid      & 92.71 & \thirdplace{98.68} & \firstplace{98.86} & 98.51 & 96.96 & 94.18 & 96.66 & 98.30 & \secondplace{98.80} & 98.09 & 98.18 \\
wbc          & \thirdplace{97.15} & 97.07 & 96.26 & 94.48 & 95.82 & 97.08 & 95.59 & \firstplace{98.21} & \secondplace{97.77} & 95.16 & 97.08 \\
WDBC         & \thirdplace{99.89} & 99.78 & 99.80 & 98.68 & 99.61 & 99.18 & 98.70 & \secondplace{99.90} & 99.77 & \firstplace{99.96} & \corethird{99.89} \\
WPBC         & 50.62 & 48.01 & 49.73 & 47.76 & 47.20 & 50.93 & 50.39 & \thirdplace{51.00} & \firstplace{55.67} & 48.30 & \coresecond{52.01} \\
yeast        & 45.71 & 44.50 & 41.82 & 45.46 & 46.40 & 44.54 & \firstplace{56.68} & 47.86 & \secondplace{49.88} & 46.56 & \corethird{48.76} \\
\midrule
\rowcolor{gray!15}
\multicolumn{12}{c}{\textbf{Out-of-Domain Target Datasets}} \\
\midrule
amazon       & 53.92 & 53.87 & 50.80 & 50.05 & 52.82 & 52.01 & 51.77 & 50.70 & \thirdplace{54.65} & \secondplace{54.69} & \corefirst{55.44} \\
backdoor     & 93.81 & 93.58 & 74.81 & 67.24 & 92.50 & \thirdplace{96.78} & 92.95 & \secondplace{97.96} & 69.70 & 95.92 & \corefirst{98.08} \\
campaign     & 64.37 & 74.07 & 73.74 & 70.91 & 77.53 & \firstplace{83.54} & 67.60 & 73.08 & \thirdplace{78.46} & 75.64 & \coresecond{79.49} \\
cover        & 91.21 & 87.96 & 74.75 & 83.35 & 86.30 & 92.01 & 94.72 & \firstplace{98.72} & \secondplace{98.50} & 96.27 & \corethird{98.44} \\
fraud        & 35.69 & 93.17 & \firstplace{94.02} & \secondplace{93.98} & \thirdplace{93.65} & 92.70 & 84.57 & 83.11 & 92.98 & 87.85 & 91.61 \\
glass        & 57.71 & 56.63 & 56.76 & 54.41 & 57.17 & 58.04 & 59.46 & 58.53 & \firstplace{89.96} & \thirdplace{66.90} & \coresecond{70.10} \\
ionosphere   & 83.44 & 94.90 & 84.19 & 89.19 & 96.05 & \thirdplace{96.76} & 95.78 & 96.71 & \secondplace{96.90} & 96.39 & \corefirst{98.05} \\
SpamBase     & 73.23 & 77.00 & \secondplace{84.74} & 77.96 & 81.80 & 74.61 & 79.73 & 83.80 & 60.25 & \firstplace{85.99} & \corethird{84.37} \\
speech       & 37.88 & 36.66 & 39.97 & 43.28 & 36.67 & 44.70 & \firstplace{56.28} & \secondplace{55.95} & \thirdplace{54.93} & 48.91 & 46.24 \\
vowels       & 84.87 & 82.21 & 59.25 & 50.15 & 81.02 & \secondplace{85.39} & 83.46 & \thirdplace{85.00} & \firstplace{92.83} & 81.51 & 84.39 \\
Wilt         & 68.81 & 75.45 & 48.16 & 38.92 & 44.19 & 74.85 & \thirdplace{78.27} & 77.90 & 75.43 & \secondplace{81.02} & \corefirst{93.36} \\
\midrule
Average      & 77.87 & 80.01 & 78.08 & 75.09 & 79.63 & 81.02 & 80.66 & \summarythird{81.76} & 81.40 & \summarysecond{83.45} & \corefirst{84.88} \\
Rank $\downarrow$ & 7.10 & 6.96 & 6.87 & 8.34 & 6.72 & 5.63 & 6.76 & 5.26 & \summarythird{4.91} & \summarysecond{4.35} & \corefirst{3.09} \\
\bottomrule
\end{tabular}
\caption{AUROC (\%) comparison between baselines and \ourmethod. Highlighted are the results ranked \firstplace{first}, \secondplace{second}, and \thirdplace{third}; colored cells indicate the rank of \ourmethod. ``Rank'' indicates the average ranking over all datasets.}
\label{tab:main_results}
\end{table*}

%% file: tables/ablation.tex
\begin{table}[t]
\centering
\small
\begin{tabular}{l|ccc|c}
\toprule
Variant & AUROC & AUPRC & F1 & Rank\\
\midrule
\ourmethod & \textbf{0.8489} & \textbf{0.6777} & \textbf{0.6459} &1 \\
\midrule
w/o Align & 0.8363 & 0.6702 & 0.6413 &2 \\
w/o Gating & 0.7999 & 0.5870 & 0.5629 &3 \\
w/o Context & 0.7856 & 0.5569 & 0.5238 &4\\
\bottomrule
\end{tabular}
\caption{Ablation results w.r.t. average performance.}
\label{tab:ablation}
\end{table}

%% file: figs/ContextRatio/contextRatio.tex
\begin{figure}[tbp]
    \centering    
    \subfigure[AUROC]{
        \label{subfig:ratio_auc}
        \includegraphics[scale=0.37]{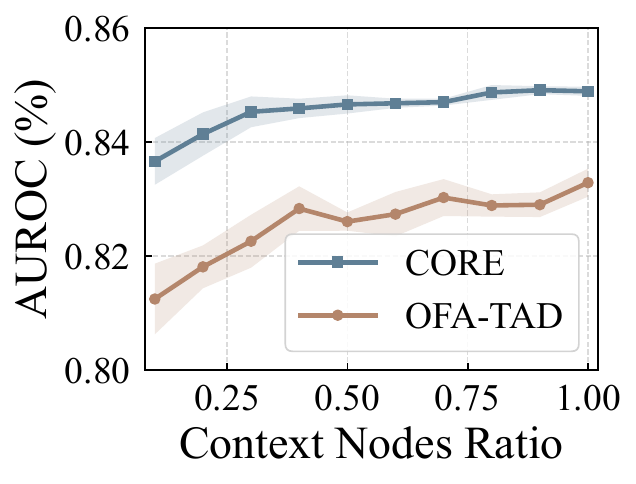}
    }\hfill    
    \subfigure[AUPRC]{
        \label{subfig:ratio_ap}
        \includegraphics[scale=0.37]{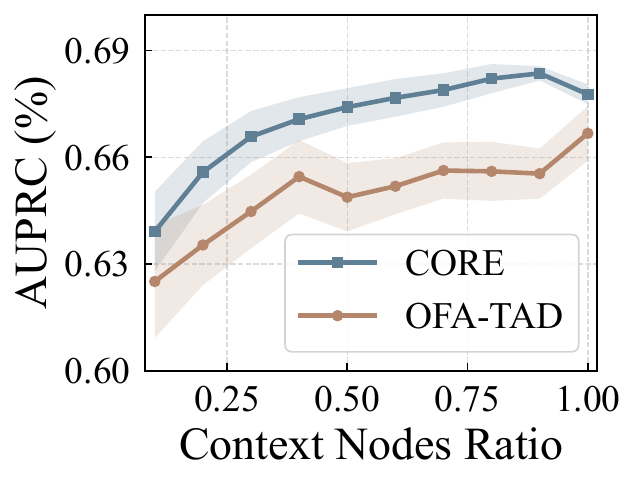}
    }
    \vspace{-2mm}
    \caption{Performance with varying context nodes.}
    \label{fig:context_ratio}
\end{figure}


%% file: figs/Efficiency/eff.tex
\begin{figure}[t]
    \centering
        \includegraphics[scale=0.51]{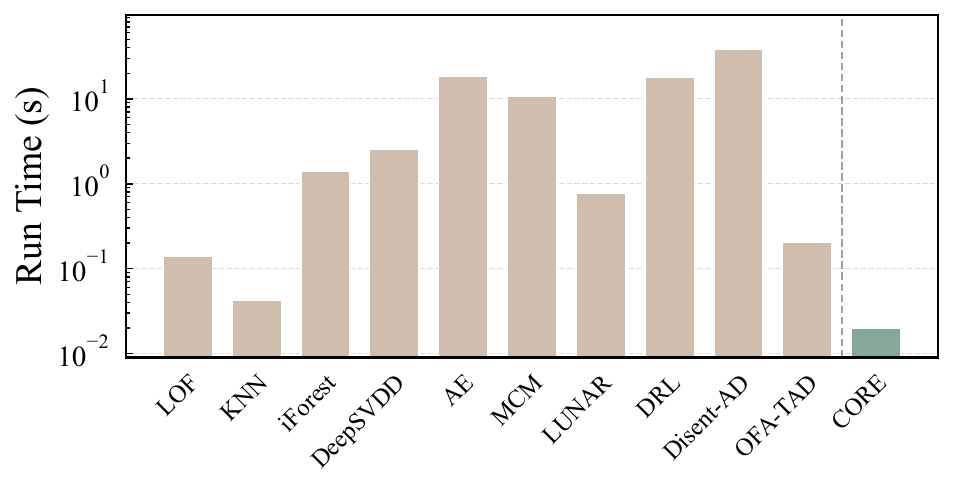}
        \vspace{-2mm}
    \caption{Overall runtime comparison in seconds.}
	\label{fig:efficiency}
    \vspace{-2mm}
\end{figure}

%% file: figs/ScalingLaw/scale.tex
\begin{figure}
\vspace{-2mm}
    \centering
        \includegraphics[scale=0.50]{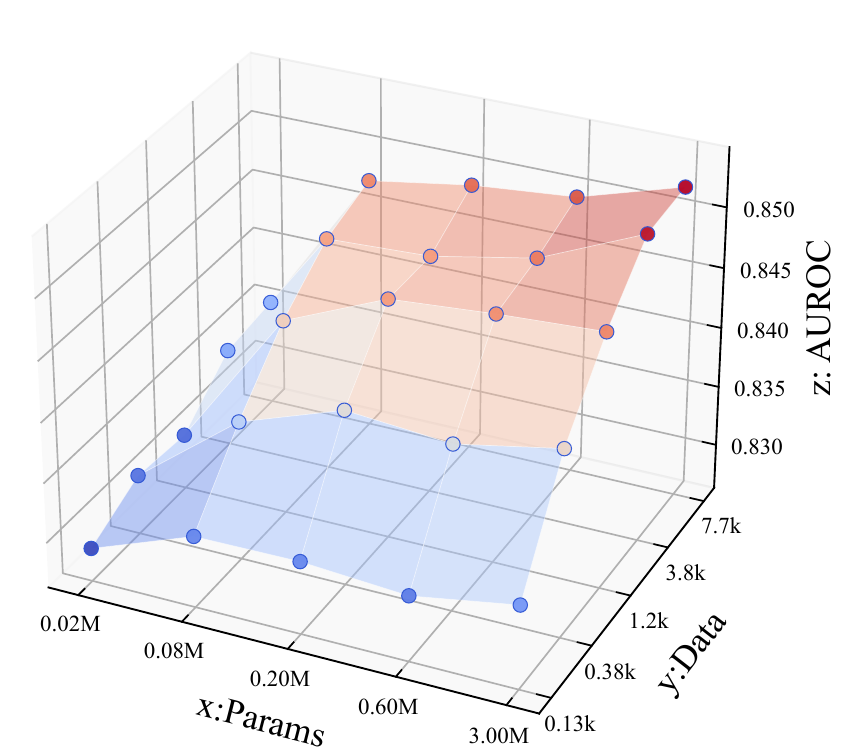}
    \caption{Neural scaling law analysis of \ourmethod.}
	\label{fig:scale}
\end{figure}

%% file: sections/6_con.tex

In this paper, we propose \ourmethod, a in-context reconstruction approach for unified TAD, aiming to detect anomalies across diverse tabular datasets without dataset-specific retraining or fine-tuning. \ourmethod first aligns heterogeneous tabular features through decorrelated feature alignment, and then reconstructs each query sample from its most relevant normal context samples through a gated aggregation mechanism. Extensive experiments on 34 real-world datasets demonstrate that \ourmethod achieves superior detection performance, efficiency, and scalability compared to existing approaches.

%% file: Appendices/sections/1_rw.tex

\noindent\textbf{Tabular Data Anomaly Detection.} Anomaly detection aims to identify abnormal instances that deviate significantly from normal patterns and has been widely studied in various real-world applications~\cite{pan2026camera,zhao2026fedcigar,tan2024influence}. Due to the broad applicability of tabular data, tabular data anomaly detection (TAD) has attracted increasing attention. Earlier studies primarily focus on classical machine learning approaches for TAD. For example, Isolation Forest~\cite{liu2008isolation} isolates anomalies through random partitioning, where anomalous samples are expected to be separated with fewer partitions. Local Outlier Factor~\cite{breunig2000lof} compares the local density of each sample with that of its neighbors and assigns higher anomaly scores to samples with lower local density. KNN-based detection~\cite{angiulli2002fast} measures the abnormality of each sample according to its distance to nearest neighbors. Although these methods are simple and efficient, they mainly rely on fixed geometric assumptions in the original feature space, which may limit their capability in handling complex tabular distributions.

Recent research highlights the effectiveness of deep learning in TAD. For instance, AE~\cite{chen2018autoencoder} employs an encoder-decoder network to reconstruct normal samples, where anomalies are assessed based on reconstruction errors. DeepSVDD~\cite{liznerski2020explainable} learns a compact representation of normal samples by mapping them into a hypersphere and uses the distance to the center as the anomaly score. LUNAR~\cite{goodge2022lunar} further exploits local neighborhood information to estimate the abnormality of each sample. Apart from these methods, another line of studies utilizes self-supervised learning to capture intrinsic regularities of tabular data. MCM~\cite{yin2024mcm} extends reconstruction-based TAD with masked cell modeling to capture feature correlations through masked reconstruction. DRL~\cite{ye2025drl} and DisentAD~\cite{ye2025disentangling} introduce disentanglement-inspired objectives to learn informative representations for anomaly detection. NeuTraLAD~\cite{qiu2021neural} utilizes learnable transformations and contrastive objectives to model characteristic patterns of normal data.


Despite their promising performance, the above methods still follow the conventional one model for one dataset paradigm, requiring dataset-specific training or tuning for each new dataset. As a result, they may struggle to generalize to unseen datasets under distribution shifts~\cite{li2026relational}. Recent studies have increasingly focused on developing more adaptable models that can generalize across diverse data scenarios~\cite{shen2026raising,qian2026dynhd,chen2025multi}, motivating the exploration of unified anomaly detection frameworks.

\noindent\textbf{Unified Anomaly Detection.} Unified Anomaly Detection aims to identify anomalies across diverse tabular datasets with a single generalist model, without dataset-specific training or fine-tuning. Inspired by the success of generalist models~\cite{brown2020language,devlin2019bert,liu2026few,li2026ofa}, recent tabular foundation models have shown strong transferable ability on general predictive tasks. However, these methods are mainly designed for general tabular prediction tasks and may struggle with TAD, where abnormality patterns are usually infrequent, irregular, and heterogeneous.

Recent studies have explored in-context prediction for unified TAD. For instance, FoMo-0D~\cite{shen2025fomo} formulates tabular outlier detection as a zero-shot prediction problem and directly identifies anomalies in unseen datasets using context samples. OUTFORMER~\cite{ding2026zero} further advances FoMo-0D by enriching synthetic detection tasks and adaptively organizing them during pre-training. Beyond the inlier-only context setting, ICLAD~\cite{wei2026iclad} unifies one-class, unsupervised, and semi-supervised TAD under an in-context learning framework. Apart from detection performance, FoMo-X~\cite{kluttermann2026fomo} studies the explainability of outlier detection foundation models by providing diagnostic signals for anomaly severity and prediction uncertainty.

Considering the inconsistent feature dimensions and semantics across tabular datasets, another line of studies constructs dataset-agnostic representations for cross-domain anomaly detection. OFA-TAD~\cite{li2026towards} extracts multi-view neighbor-distance profiles as transferable anomaly cues and fuses view-specific anomaly evidence with a Mixture-of-Experts network. To support classification-based training under the one-class setting, OFA-TAD further synthesizes pseudo-anomalies through multiple negative sampling strategies. Meanwhile, recent GNN-based anomaly detection methods have demonstrated promising performance by effectively capturing structural dependencies and relational information among samples~\cite{liu2026few,liu2026rethinking,pan2026correcting,zheng2026unsupervised}. Motivated by this, UniOD converts each dataset into kernel-based similarity graphs and employs SVD-derived node embeddings to obtain consistent node features, thereby framing outlier detection as a node classification task.

While existing unified TAD methods have demonstrated promising results, they mainly follow a classification-based paradigm, requiring anomaly supervision from labeled anomalies or synthesized pseudo-anomalies. However, real anomaly labels are usually scarce, and synthetic anomalies may deviate from realistic abnormal patterns. Therefore, how to achieve unified TAD without such anomaly supervision still remains under-explored.

%% file: Appendices/sections/2_algo.tex
\input{Appendices/algorithm/algo_align}

\input{Appendices/algorithm/algo_train}
\input{Appendices/algorithm/algo_infer}
\subsection{Algorithmic Description}
The algorithmic description of the feature alignment in \ourmethod, the training process of \ourmethod, and inference process of \ourmethod are summarized in Algo.~\ref{alg:align}, Algo.~\ref{alg:train}, and Algo.~\ref{alg:inference}, respectively.

\subsection{Complexity Analysis}
The time complexity of \ourmethod primarily consists of two main components: decorrelated feature alignment and in-context gated reconstruction. For decorrelated feature alignment, the overall complexity is $O(nd^2+dd_u^2+nd_u)$, where $n$, $d$, and $d_u$ denote the number of samples involved in feature alignment, the original feature dimension, and the unified feature dimension, respectively. Here, the first term is used for computing inter-feature correlations, while the second and third terms are used for greedy feature selection and feature rearrangement, respectively.

For in-context gated reconstruction, let $n_c$ and $n_q$ denote the number of context and query samples, respectively, $h$ denote the hidden dimension, $p$ denote the metric embedding dimension, and $K$ denote the number of selected context samples. The model inference is divided into three main parts: multi-head encoding, context weighting, and gated reconstruction. The complexity of multi-head encoding is $O((n_c+n_q)(d_uh+h^2+hp)+n_chd_u)$, where the first term is used for the shared backbone and metric head, and the second term is used for the value and gate heads of context samples. The context weighting mainly involves query-context distance calculation and nearest neighbor retrieval, with time complexity $O(n_qn_cp+n_qn_c\log K)$. The gated fusion and weighted reconstruction require $O(n_qKd_u)$.

Therefore, the overall inference complexity of \ourmethod is $O(nd^2+dd_u^2+nd_u+(n_c+n_q)(d_uh+h^2+hp)+n_chd_u+n_qn_c(p+\log K)+n_qKd_u)$, where the first three terms correspond to decorrelated feature alignment, the terms $(n_c+n_q)(d_uh+h^2+hp)+n_chd_u$ are used for multi-head encoding, the term $n_qn_c(p+\log K)$ is used for context weighting, and the term $n_qKd_u$ is used for gated fusion and weighted reconstruction.

%% file: Appendices/algorithm/algo_align.tex
\begin{algorithm}[tb]
\caption{Decorrelated Feature Alignment}
\label{alg:align}
\textbf{Input}: Feature matrix $\mathbf{X}\in\mathbb{R}^{n\times d}$.\\
\textbf{Parameters}: Unified dimension $d_u$.\\
\textbf{Output}: Aligned feature matrix $\mathbf{X}'$.
\begin{algorithmic}[1]
\STATE Calculate feature redundancy $\rho_j^{\operatorname{max}}$ for each feature via Eq.~(1)
\STATE Calculate relevance score $s_j$ for each feature via Eq.~(2)
\STATE Initialize selected feature set $\mathcal{S}$ with the feature of the highest $s_j$
\WHILE{$|\mathcal{S}| < d_u$}
    \STATE Calculate candidate score $s_j^{*}$ for each unselected feature via Eq.~(3)
    \STATE Select feature $j^{\star}$ and update $\mathcal{S}$ via Eq.~(4)
\ENDWHILE
\STATE $\mathbf{X}' \leftarrow$ Rearrange features of $\mathbf{X}$ according to the selected order $\mathcal{S}$
\STATE \textbf{return} $\mathbf{X}'$
\end{algorithmic}
\end{algorithm}


%% file: Appendices/algorithm/algo_train.tex
\begin{algorithm}[tb]
\caption{Training Procedure of \ourmethod}
\label{alg:train}
\textbf{Input}: Source dataset collection $\mathcal{D}_{\mathrm{source}}$.\\
\textbf{Parameters}: Unified dimension $d_u$; context size $n_c$; neighbors $K$; epochs $T$; temperature $\tau$.\\
\textbf{Output}: Trained model parameters $\Theta$.
\begin{algorithmic}[1]
\STATE Initialize model parameters $\Theta$ of $f_{\theta}$, $g_{\mathrm{m}}$, $g_{\mathrm{v}}$, and $g_{\mathrm{g}}$
\FOR{each source dataset $\mathcal{D}_j \in \mathcal{D}_{\mathrm{source}}$}
    \STATE Obtain training feature matrix $\mathbf{X}_{j,\mathrm{train}}$ from $\mathcal{D}_{j,\mathrm{train}}$
    \STATE $\mathbf{X}'_{j,\mathrm{train}} \leftarrow$ Align $\mathbf{X}_{j,\mathrm{train}}$ via Algo.~\ref{alg:align}
\ENDFOR
\FOR{$t=1$ to $T$}
    \FOR{each aligned source feature matrix $\mathbf{X}'_{j,\mathrm{train}}$}
        \STATE Randomly split $\mathbf{X}'_{j,\mathrm{train}}$ into context set $\mathbf{X}'_c$ and query set $\mathbf{X}'_q$
        \STATE $\mathbf{H}_q,\mathbf{H}_c \leftarrow$ Encode $\mathbf{X}'_q$ and $\mathbf{X}'_c$ via Eq.~(7)
        \STATE $\mathbf{M}_q \leftarrow g_{\mathrm{m}}(\mathbf{H}_q)$
        \STATE $\mathbf{M}_c,\mathbf{V}_c,\mathbf{G}_c \leftarrow$ Calculate context representations via Eq.~(8)
        \STATE Retrieve $K$ nearest context samples for each query based on $\mathbf{M}_q$ and $\mathbf{M}_c$
        \STATE Calculate context weights $\{w_{i,k}\}_{k=1}^{K}$ via Eq.~(9)
        \STATE Calculate fused context representations $\{\mathbf{z}_{i,k}\}_{k=1}^{K}$ via Eq.~(10)
        \STATE Reconstruct each query sample $\hat{\mathbf{x}}_i$ via Eq.~(11)
        \STATE Calculate reconstruction loss $\mathcal{L}=\frac{1}{|\mathbf{X}'_q|}\sum_{\mathbf{x}'_i\in\mathbf{X}'_q}\|\mathbf{x}'_i-\hat{\mathbf{x}}_i\|_2^2$
        \STATE Update $\Theta$ by gradient descent
    \ENDFOR
\ENDFOR
\end{algorithmic}
\end{algorithm}

%% file: Appendices/algorithm/algo_infer.tex
\begin{algorithm}[tb]
\caption{Inference Procedure of \ourmethod}
\label{alg:inference}
\textbf{Input}: Target dataset $\mathcal{D}_j \in \mathcal{D}_{\mathrm{target}}$.\\
\textbf{Parameters}: Trained model parameters $\Theta$; unified dimension $d_u$; neighbors $K$; temperature $\tau$.\\
\textbf{Output}: Anomaly scores $\{a(i)\}$ for samples in $\mathcal{D}_{j,\mathrm{test}}$.
\begin{algorithmic}[1]
\STATE Obtain feature matrices $\mathbf{X}_{j,\mathrm{train}}$ and $\mathbf{X}_{j,\mathrm{test}}$ from $\mathcal{D}_{j,\mathrm{train}}$ and $\mathcal{D}_{j,\mathrm{test}}$
\STATE $\mathbf{X}'_c \leftarrow$ Align $\mathbf{X}_{j,\mathrm{train}}$ via Algo.~\ref{alg:align}
\STATE $\mathbf{X}'_q \leftarrow$ Transform $\mathbf{X}_{j,\mathrm{test}}$ using the same fitted alignment
\STATE $\mathbf{H}_c \leftarrow f_{\theta}(\mathbf{X}'_c)$
\STATE $\mathbf{M}_c,\mathbf{V}_c,\mathbf{G}_c \leftarrow$ Calculate context representations via Eq.~(8)
\STATE $\mathbf{H}_q \leftarrow f_{\theta}(\mathbf{X}'_q)$; $\mathbf{M}_q \leftarrow g_{\mathrm{m}}(\mathbf{H}_q)$
\FOR{each query sample $\mathbf{x}'_i \in \mathbf{X}'_q$}
    \STATE Retrieve $K$ nearest context samples from $\mathbf{X}'_c$ based on $\mathbf{M}_q$ and $\mathbf{M}_c$
    \STATE Calculate context weights $\{w_{i,k}\}_{k=1}^{K}$ via Eq.~(9)
    \STATE Calculate fused context representations $\{\mathbf{z}_{i,k}\}_{k=1}^{K}$ via Eq.~(10)
    \STATE Reconstruct query sample $\hat{\mathbf{x}}_i$ via Eq.~(11)
    \STATE Calculate anomaly score $a(i)$ via Eq.~(12)
\ENDFOR
\STATE \textbf{return} $\{a(i)\}$ as anomaly scores for query samples
\end{algorithmic}
\end{algorithm}

%% file: Appendices/sections/3_detail.tex
\input{Appendices/tables/dsets}

\subsection{Detail of Datasets} \label{app:datasets}
The detailed statistics of the datasets are reported in Table~\ref{tab:detail_dsets}. Following OFA-TAD, we adopt a category-aware split to evaluate the unified generalization ability of \ourmethod. Specifically, 7 datasets are used as source datasets for training, and 34 datasets are used as target datasets for evaluation, including 23 in-domain datasets and 11 out-of-domain datasets.

Each dataset is associated with a semantic category. For categories containing at least three datasets, we select one dataset as the source dataset and use the remaining datasets from the same category as in-domain target datasets. For the Healthcare category, which contains more datasets than other categories, two datasets are selected as source datasets. Categories with fewer than three datasets are regarded as unseen categories, and their datasets are used only for out-of-domain evaluation. In Table~\ref{tab:detail_dsets}, the ``Train'' and ``Test'' columns denote whether each dataset is used for source training or target evaluation, respectively. We also report the number of samples, feature dimension, number of anomalies, and anomaly ratio for each dataset.

\subsection{Description of Baselines}\label{app:baseline}
In our evaluation, we provide a comprehensive comparison of \ourmethod with classical machine learning methods, deep learning-based methods, and unified TAD methods. For classical and deep TAD baselines, we follow the standard dataset-specific evaluation protocol, where each model is fitted on the target training split and evaluated on the corresponding target test split. For unified methods, the model is trained once on source datasets and directly applied to target datasets without target-specific retraining. For classical machine learning methods, we consider three widely used TAD baselines:
\begin{itemize}
    \item \textbf{IForest}~\cite{liu2008isolation} isolates anomalies through random partitioning. It assumes that anomalous samples are easier to isolate and therefore require fewer partitioning steps than normal samples.
    
    \item \textbf{LOF}~\cite{breunig2000lof} measures the local density of each sample and compares it with the densities of its neighbors. Samples with substantially lower local density are assigned higher anomaly scores.
    
    \item \textbf{KNN}~\cite{angiulli2002fast} estimates the abnormality of each sample based on its distances to nearest neighbors, where samples far from their neighbors are more likely to be detected as anomalies.
\end{itemize}

\noindent For deep learning-based methods, we compare with several representative TAD models based on reconstruction, one-class learning, neighborhood modeling, and representation learning:
\begin{itemize}
    \item \textbf{AE}~\cite{chen2018autoencoder} employs an encoder-decoder network to reconstruct normal samples and uses reconstruction errors as anomaly scores.
    
    \item \textbf{DeepSVDD}~\cite{liznerski2020explainable} maps normal samples into a compact hypersphere in the latent space and identifies anomalies according to their distances to the hypersphere center.
    
    \item \textbf{LUNAR}~\cite{goodge2022lunar} exploits local neighborhood information for anomaly detection by modeling the relationship between each sample and its nearest neighbors.
    
    \item \textbf{MCM}~\cite{yin2024mcm} introduces masked cell modeling for TAD, which captures feature correlations by reconstructing masked feature values.
    
    \item \textbf{DRL}~\cite{ye2025drl} learns anomaly-aware representations through a deep representation learning framework, aiming to better capture intrinsic normal patterns for anomaly detection.
    
    \item \textbf{DisentAD}~\cite{ye2025disentangling} utilizes disentanglement-inspired representation learning to separate different latent factors in tabular data for anomaly detection.
\end{itemize}

\noindent For the unified TAD baseline, we include the recent generalist model:
\begin{itemize}
    \item \textbf{OFA-TAD}~\cite{li2026towards} extracts multi-view neighbor-distance profiles as transferable anomaly cues and fuses view-specific anomaly evidence with a Mixture-of-Experts network. It is trained once on source datasets and transferred to unseen target datasets under the unified protocol.
\end{itemize}

\subsection{Details of Implementation} \label{app:implementation}

\noindent\textbf{Hyperparameter Settings.} We select some key hyper-parameters of \ourmethod through random search within specified grids. Specifically, the random search was performed within the following search space:
\begin{itemize}
    \item Hidden layer dimension: $\{64, 128, 256, 512\}$.
    \item Number of MLP layers: $\{1, 2, 3, 4\}$.
    \item Metric embedding dimension $p$: $\{32, 64, 128, 256\}$.
    \item Number of selected context samples $K$: $\{3, 5, 10, 20\}$.
    \item Learning rate: floats between $10^{-5}$ and $10^{-2}$.
    \item Weight decay: floats between $10^{-6}$ and $10^{-3}$.
\end{itemize}

\noindent\textbf{Evaluation Metrics.} Following OFA-TAD~\cite{li2026towards}, we employ three widely used evaluation metrics for TAD, including the area under the receiver operating characteristic curve (AUROC), the area under the precision-recall curve (AUPRC), and F1-Score. A higher value indicates better detection performance. We report the average results with standard deviations across five trials.

\noindent\textbf{Environment.} \ourmethod is implemented with Python 3.10, CUDA 12.1, PyTorch 2.1.2, torchvision 0.16.2, torchaudio 2.1.2, and NumPy 1.26.4.

\noindent\textbf{Hardware Configuration.} All experiments were conducted on a Linux server equipped with an Intel(R) Core(TM) i5-13600K CPU and an NVIDIA GeForce RTX 3090 GPU.

%% file: Appendices/tables/dsets.tex
\begin{table*}[!t]
\centering
\small
\setlength{\tabcolsep}{3.2pt}
\begin{tabularx}{\textwidth}{@{}
>{\raggedright\arraybackslash}X
>{\raggedright\arraybackslash}X
c
c
r
r
r
r
@{}}
\toprule
Dataset & Category & Train & Test & \# Samples & \# Features & \# Anomaly & Anomaly (\%) \\
\midrule
\rowcolor{gray!10}
\multicolumn{8}{c}{\textbf{In-domain datasets}} \\
\midrule

\rowcolor{gray!15}
satellite & Astronautics & $\surd$ & $-$ & 6435 & 36 & 2036 & 31.64 \\
\rowcolor{gray!15}
satimage-2 & Astronautics & $-$ & $\surd$ & 5803 & 36 & 71 & 1.22 \\
\rowcolor{gray!15}
shuttle & Astronautics & $-$ & $\surd$ & 49097 & 9 & 3511 & 7.15 \\

vertebral & Biology & $\surd$ & $-$ & 240 & 6 & 30 & 12.50 \\
yeast & Biology & $-$ & $\surd$ & 1484 & 8 & 507 & 34.16 \\
abalone & Biology & $-$ & $\surd$ & 4177 & 7 & 2081 & 49.82 \\

\rowcolor{gray!15}
annthyroid & Healthcare & $\surd$ & $-$ & 7200 & 6 & 534 & 7.42 \\
\rowcolor{gray!15}
breastw & Healthcare & $-$ & $\surd$ & 683 & 9 & 239 & 34.99 \\
\rowcolor{gray!15}
cardio & Healthcare & $-$ & $\surd$ & 1831 & 21 & 176 & 9.61 \\
\rowcolor{gray!15}
Cardiotocography & Healthcare & $\surd$ & $-$ & 2114 & 21 & 466 & 22.04 \\
\rowcolor{gray!15}
Hepatitis & Healthcare & $-$ & $\surd$ & 80 & 19 & 13 & 16.25 \\
\rowcolor{gray!15}
lympho & Healthcare & $-$ & $\surd$ & 148 & 18 & 6 & 4.05 \\
\rowcolor{gray!15}
mammography & Healthcare & $-$ & $\surd$ & 11183 & 6 & 260 & 2.32 \\
\rowcolor{gray!15}
Pima & Healthcare & $-$ & $\surd$ & 768 & 8 & 268 & 34.90 \\
\rowcolor{gray!15}
thyroid & Healthcare & $-$ & $\surd$ & 3772 & 6 & 93 & 2.47 \\
\rowcolor{gray!15}
WDBC & Healthcare & $-$ & $\surd$ & 367 & 30 & 10 & 2.72 \\
\rowcolor{gray!15}
WPBC & Healthcare & $-$ & $\surd$ & 198 & 33 & 47 & 23.74 \\
\rowcolor{gray!15}
wbc & Healthcare & $-$ & $\surd$ & 378 & 30 & 21 & 5.60 \\
\rowcolor{gray!15}
arrhythmia & Healthcare & $-$ & $\surd$ & 452 & 274 & 66 & 15.00 \\
\rowcolor{gray!15}
Parkinson & Healthcare & $-$ & $\surd$ & 195 & 22 & 147 & 75.38 \\

mnist & Image & $-$ & $\surd$ & 7603 & 100 & 700 & 9.21 \\
optdigits & Image & $-$ & $\surd$ & 5216 & 64 & 150 & 2.88 \\
pendigits & Image & $-$ & $\surd$ & 6870 & 16 & 156 & 2.27 \\
imgseg & Image & $\surd$ & $-$ & 2310 & 18 & 990 & 42.86 \\

\rowcolor{gray!15}
fault & Phys./Chem. & $-$ & $\surd$ & 1941 & 27 & 673 & 34.67 \\
\rowcolor{gray!15}
musk & Phys./Chem. & $-$ & $\surd$ & 3062 & 166 & 97 & 3.17 \\
\rowcolor{gray!15}
wine & Phys./Chem. & $\surd$ & $-$ & 129 & 13 & 10 & 7.75 \\

census & Sociology & $-$ & $\surd$ & 299285 & 500 & 18568 & 6.20 \\
comm.and.crime & Sociology & $\surd$ & $-$ & 1994 & 101 & 993 & 49.80 \\
donors & Sociology & $-$ & $\surd$ & 619326 & 10 & 36710 & 5.93 \\

\midrule
\rowcolor{gray!10}
\multicolumn{8}{c}{\textbf{Out-of-domain datasets}} \\
\midrule

\rowcolor{gray!15}
cover & Botany & $-$ & $\surd$ & 286048 & 10 & 2747 & 0.96 \\
\rowcolor{gray!15}
Wilt & Botany & $-$ & $\surd$ & 4819 & 5 & 257 & 5.33 \\

SpamBase & Document & $-$ & $\surd$ & 4207 & 57 & 1679 & 39.91 \\

\rowcolor{gray!15}
campaign & Finance & $-$ & $\surd$ & 41188 & 62 & 4640 & 11.27 \\
\rowcolor{gray!15}
fraud & Finance & $-$ & $\surd$ & 284807 & 29 & 492 & 0.17 \\

glass & Forensic & $-$ & $\surd$ & 214 & 7 & 9 & 4.21 \\

\rowcolor{gray!15}
speech & Linguistics & $-$ & $\surd$ & 3686 & 400 & 61 & 1.65 \\
\rowcolor{gray!15}
vowels & Linguistics & $-$ & $\surd$ & 1456 & 12 & 50 & 3.43 \\

backdoor & Network & $-$ & $\surd$ & 95329 & 196 & 2329 & 2.44 \\

\rowcolor{gray!15}
amazon & NLP & $-$ & $\surd$ & 10000 & 768 & 500 & 5.00 \\

Ionosphere & Oryctognosy & $-$ & $\surd$ & 351 & 32 & 126 & 35.90 \\
\bottomrule
\end{tabularx}
\caption{Dataset statistics under the one-for-all evaluation protocol. ``Train'' indicates source datasets used for model training, and ``Test'' indicates target datasets used for evaluation.}
\label{tab:detail_dsets}
\end{table*}

%% file: Appendices/sections/4_SuppleExp.tex
\input{Appendices/tables/main_auprc}
\input{Appendices/tables/main_f1}
\subsection{Comparison in More Metrics}\label{app:performance}
Table~\ref{tab:auprc_results} presents the AUPRC results. \ourmethod achieves the best average AUPRC of $68.47\%$ and the best average rank of $3.09$, outperforming OFA-TAD by $2.18$ points and the strongest dataset-specific baseline MCM by $5.88$ points. The best-performing method varies across datasets, indicating that different baselines capture different dataset-specific anomaly patterns. In contrast, \ourmethod shows more consistent performance under the unified tad setting.

\noindent Table~\ref{tab:f1_results} reports the F1-Score results. \ourmethod again ranks first on average, achieving $65.31\%$ F1-Score and an average rank of $3.88$. It surpasses OFA-TAD by $1.79$ points and the strongest dataset-specific baseline LUNAR by $4.10$ points. Although some datasets are threshold-sensitive and favor different methods, \ourmethod achieves the best overall F1-Score without using labeled anomalies or synthesized pseudo-anomalies, demonstrating the effectiveness of in-context reconstruction for unified TAD.

\subsection{Effectiveness of Context Ratio} \label{app:context}
In the main paper, we report the average results over all 34 target datasets. To further analyze the influence of context ratio, we select 12 representative datasets and show their AUROC and AUPRC curves in Fig.~\ref{fig:context_ratio_auc_datasets} and Fig.~\ref{fig:context_ratio_ap_datasets}. From the figure, we observe that in most cases the performance of \ourmethod increases or remains stable with the involvement of more context samples. For example, \ourmethod achieves clear improvements on Wilt, ionosphere, glass, and cover, indicating that more context samples can provide richer normal patterns for in-context reconstruction.

In addition, the curves show that the effect of context ratio is dataset-dependent. On some datasets, the performance becomes stable with a moderate context ratio, while on others the curves fluctuate when more context samples are included. This suggests that different target datasets contain heterogeneous normal patterns and local density structures. Compared with OFA-TAD, \ourmethod generally shows stronger AUPRC performance on the selected datasets, demonstrating that reconstructing queries from relevant context samples can better utilize target-domain normal information under different context ratios.

\subsection{Efficiency Analysis} \label{app:efficiency}
In the main paper, we report the efficiency comparison on the Wilt dataset. To further verify the runtime efficiency of \ourmethod across different target datasets, we select 12 representative datasets and visualize the total runtime of all compared methods in Fig.~\ref{fig:ef_deatil}. These datasets include arrhythmia, mammography, backdoor, ionosphere, Wilt, Hepatitis, Parkinson, WPBC, campaign, glass, cover, and fraud, covering both small-scale and relatively large-scale target datasets.

\noindent From the figure, we observe that \ourmethod consistently achieves highly competitive runtime efficiency across the selected datasets. Different from most deep learning-based baselines, \ourmethod does not require target-specific training or fine-tuning, and thus its training time on the target dataset is zero. This property substantially reduces the overall computation cost under the unified TAD setting. In contrast, training-based methods such as AE, DeepSVDD, MCM, DRL, LUNAR, and DisentAD usually require additional optimization on each target dataset, leading to much larger total runtime, especially on relatively large datasets such as campaign, cover, and fraud.

\noindent Compared with OFA-TAD, \ourmethod also shows clear efficiency advantages on all selected datasets. For example, on backdoor, campaign, cover, and fraud, \ourmethod requires much less total runtime than OFA-TAD, indicating that the proposed in-context gated reconstruction framework can achieve efficient target-domain inference without introducing heavy computational overhead. Moreover, \ourmethod is also comparable to, and in many cases faster than, classical methods such as LOF, KNN, and iForest. Although some extremely lightweight classical methods may have slightly lower runtime on a few small datasets, they require dataset-specific fitting and generally show weaker anomaly detection performance. Overall, these results demonstrate that \ourmethod achieves a favorable balance between detection effectiveness and runtime efficiency.

\input{Appendices/figs/context/detail_ratio_auc}
\input{Appendices/figs/context/detail_ratio_ap}
\input{Appendices/figs/Efficiency/detail_ef}

%% file: Appendices/tables/main_auprc.tex
\begin{table*}[!t]
\centering
\scriptsize
\setlength{\tabcolsep}{4.0pt}
\renewcommand{\arraystretch}{0.65}
\setlength{\aboverulesep}{0.15ex}
\setlength{\belowrulesep}{0.15ex}
\resizebox{\textwidth}{!}{%
\begin{tabular}{l|ccc|cccccc|c|>{\columncolor{corecolbg}}c}
\toprule
Dataset & LOF & KNN & iForest & DSVDD & AE & MCM & LUNAR & DRL & DisentAD & OFA-TAD & \ourmethod \\
\midrule
\rowcolor{gray!15}
\multicolumn{12}{c}{\textbf{In-Domain Target Datasets}} \\
\midrule
abalone      & 82.73 & 87.99 & 84.96 & 80.56 & \thirdplace{88.56} & 86.02 & 88.39 & \secondplace{88.62} & 87.00 & \firstplace{89.03} & 87.42 \\
arrhythmia   & \thirdplace{61.14} & 60.34 & 57.69 & 50.91 & 39.99 & 58.32 & 56.77 & 56.76 & 53.96 & \secondplace{62.12} & \corefirst{66.02} \\
breastw      & 95.32 & \thirdplace{99.62} & \secondplace{99.73} & 98.94 & 96.70 & \firstplace{99.77} & 96.89 & 99.45 & 99.61 & 97.40 & 99.32 \\
cardio       & 75.99 & 76.63 & 78.86 & \firstplace{84.25} & \secondplace{81.94} & 55.93 & 78.26 & 81.17 & 80.85 & 80.58 & \corethird{81.75} \\
census       & 11.75 & 16.26 & 14.51 & \thirdplace{19.05} & \secondplace{20.74} & 16.50 & 16.54 & 15.00 & \firstplace{40.03} & 17.48 & 17.85 \\
donors       & 77.73 & \thirdplace{97.25} & 43.45 & 26.95 & 47.52 & 92.93 & 82.83 & 41.57 & 60.24 & \secondplace{99.29} & \corefirst{99.94} \\
fault        & 50.44 & 61.98 & 60.02 & 55.12 & 59.53 & \firstplace{65.22} & 57.25 & \secondplace{63.09} & 62.59 & 61.50 & \corethird{62.67} \\
Hepatitis    & 42.89 & 30.63 & 42.89 & \firstplace{54.44} & 39.85 & 45.15 & 31.15 & 43.44 & 47.67 & \thirdplace{48.46} & \coresecond{54.06} \\
lympho       & 56.46 & 46.28 & \thirdplace{97.38} & \secondplace{98.67} & 40.61 & 92.46 & 85.59 & 87.84 & 57.09 & 91.82 & \corefirst{100.00} \\
mammography  & 35.00 & 38.93 & 41.25 & \thirdplace{43.17} & 38.99 & 43.03 & 26.19 & 38.13 & 41.50 & \secondplace{53.95} & \corefirst{54.38} \\
mnist        & \thirdplace{83.80} & 76.93 & 55.15 & 57.01 & 69.85 & \secondplace{86.32} & 67.10 & \firstplace{86.58} & 30.30 & 81.04 & 77.86 \\
musk         & \firstplace{100.00} & \firstplace{100.00} & 77.20 & 89.27 & \firstplace{100.00} & \thirdplace{99.17} & 92.79 & \secondplace{99.86} & 78.08 & \firstplace{100.00} & \corefirst{100.00} \\
optdigits    & \secondplace{81.72} & 48.83 & 14.15 & 5.91 & 16.08 & 79.77 & 58.25 & 25.55 & \firstplace{84.24} & \thirdplace{81.62} & 81.57 \\
Parkinson    & 92.99 & 82.02 & \firstplace{96.07} & 92.61 & 92.92 & 79.27 & 83.36 & 92.11 & 92.09 & \thirdplace{93.48} & \coresecond{94.83} \\
pendigits    & 65.49 & 72.36 & 50.19 & 19.25 & 53.29 & 70.68 & 85.08 & 42.69 & \thirdplace{89.96} & \firstplace{97.58} & \coresecond{97.43} \\
pima         & 68.56 & \thirdplace{71.81} & \secondplace{71.94} & 67.07 & 71.22 & 70.62 & 68.87 & \firstplace{72.10} & 68.56 & 70.37 & 71.56 \\
satimage-2   & 88.46 & \thirdplace{96.69} & 94.61 & 83.87 & \firstplace{97.79} & \secondplace{97.74} & 61.79 & 86.02 & 57.36 & 96.10 & 95.70 \\
shuttle      & 94.58 & 92.25 & 98.62 & 96.22 & 96.83 & 96.83 & 81.38 & 96.93 & \thirdplace{99.58} & \secondplace{99.61} & \corefirst{99.80} \\
thyroid      & 60.55 & \secondplace{80.94} & 75.06 & \thirdplace{80.38} & 72.64 & 78.17 & 67.78 & 78.67 & \firstplace{86.85} & 80.26 & 77.71 \\
wbc          & \secondplace{85.73} & 84.38 & 83.17 & 79.10 & 80.23 & 84.13 & 76.38 & \firstplace{91.14} & 81.20 & 80.92 & \corethird{85.40} \\
WDBC         & \thirdplace{98.33} & 95.73 & 96.77 & 87.23 & 94.85 & 94.62 & 92.50 & \secondplace{98.43} & 96.05 & \firstplace{99.33} & 98.07 \\
WPBC         & 41.19 & 40.37 & 37.99 & 38.72 & 38.48 & \secondplace{42.99} & 39.84 & 41.01 & \firstplace{43.14} & 40.20 & \corethird{42.07} \\
yeast        & 48.66 & 48.21 & 46.97 & 48.58 & 47.99 & 47.66 & \firstplace{55.04} & 50.30 & \thirdplace{51.25} & 49.12 & \coresecond{51.35} \\
\midrule
\rowcolor{gray!15}
\multicolumn{12}{c}{\textbf{Out-of-Domain Target Datasets}} \\
\midrule
amazon       & 10.10 & 9.99 & 9.47 & 9.52 & 9.87 & 9.74 & \firstplace{19.36} & 9.61 & \secondplace{11.82} & 10.28 & \corethird{10.81} \\
backdoor     & 37.51 & 47.99 & 9.43 & 10.08 & \secondplace{85.58} & 61.54 & 62.93 & \firstplace{86.19} & 12.30 & \thirdplace{73.00} & 72.25 \\
campaign     & 27.68 & 44.67 & \thirdplace{46.14} & 42.41 & \thirdplace{46.14} & \firstplace{55.70} & 42.41 & 45.90 & 43.99 & 45.15 & \coresecond{49.27} \\
cover        & 15.98 & 10.45 & 5.60 & 5.95 & 6.61 & 15.10 & 44.87 & \firstplace{82.31} & \thirdplace{71.78} & 53.70 & \coresecond{74.01} \\
fraud        & 0.27 & 25.35 & 23.73 & 18.19 & 27.77 & \secondplace{53.06} & \thirdplace{39.81} & 23.09 & \firstplace{61.42} & 38.70 & 30.08 \\
glass        & 9.52 & 9.31 & 9.56 & 9.04 & 9.46 & 14.47 & 10.10 & 10.08 & \firstplace{48.90} & \thirdplace{17.68} & \coresecond{21.78} \\
ionosphere   & 86.07 & 95.90 & 85.59 & 90.66 & 96.95 & \thirdplace{97.72} & 97.00 & \secondplace{97.74} & 97.65 & 97.42 & \corefirst{98.42} \\
SpamBase     & 72.71 & 81.35 & \secondplace{87.60} & 79.71 & 82.00 & 78.33 & 81.60 & \thirdplace{84.55} & 65.24 & \firstplace{89.24} & 83.94 \\
speech       & 2.95 & 2.73 & 3.18 & 2.90 & 2.72 & 3.24 & \secondplace{5.78} & \thirdplace{4.72} & \firstplace{6.85} & 3.68 & 3.30 \\
vowels       & 30.56 & 30.21 & 11.56 & 10.52 & 30.18 & \secondplace{36.68} & 31.79 & 31.75 & \firstplace{51.26} & 32.00 & \corethird{35.37} \\
Wilt         & 15.74 & 17.46 & 8.81 & 7.78 & 8.25 & 19.09 & \secondplace{31.68} & \thirdplace{26.84} & 18.76 & 21.65 & \corefirst{52.14} \\
\midrule
Average      & 56.14 & 58.29 & 53.51 & 51.30 & 55.65 & \summarythird{62.59} & 59.33 & 61.15 & 61.15 & \summarysecond{66.29} & \corefirst{68.47} \\
Rank $\downarrow$ & 7.38 & 6.62 & 7.38 & 8.28 & 7.03 & 5.31 & 6.75 & \summarythird{5.06} & 5.16 & \summarysecond{3.94} & \corefirst{3.09} \\
\bottomrule
\end{tabular}
}
\caption{AUPRC (\%) comparison between baselines and \ourmethod. Highlighted are the results ranked \firstplace{first}, \secondplace{second}, and \thirdplace{third}; colored cells indicate the rank of \ourmethod. ``Rank'' indicates the average ranking over all datasets.}
\label{tab:auprc_results}
\end{table*}

%% file: Appendices/tables/main_f1.tex
\begin{table*}[!t]
\centering
\scriptsize
\setlength{\tabcolsep}{4.0pt}
\renewcommand{\arraystretch}{0.65}
\setlength{\aboverulesep}{0.15ex}
\setlength{\belowrulesep}{0.15ex}
\resizebox{\textwidth}{!}{%
\begin{tabular}{l|ccc|cccccc|c|>{\columncolor{corecolbg}}c}
\toprule
Dataset & LOF & KNN & iForest & DSVDD & AE & MCM & LUNAR & DRL & DisentAD & OFA-TAD & \ourmethod \\
\midrule
\rowcolor{gray!15}
\multicolumn{12}{c}{\textbf{In-Domain Target Datasets}} \\
\midrule
abalone      & 77.85 & 81.07 & 77.10 & 73.95 & \thirdplace{81.40} & 77.20 & \secondplace{81.54} & 81.31 & 79.57 & \firstplace{82.29} & 80.27 \\
arrhythmia   & 57.58 & \thirdplace{59.09} & \thirdplace{59.09} & 51.21 & 48.48 & 57.58 & 57.80 & 50.61 & 51.52 & \firstplace{60.61} & \coresecond{59.70} \\
breastw      & 94.56 & \thirdplace{97.49} & \firstplace{98.16} & 96.82 & 94.14 & \thirdplace{97.49} & 95.26 & \secondplace{97.82} & 97.07 & 94.39 & 96.07 \\
cardio       & 67.05 & 68.18 & 67.08 & \firstplace{74.77} & 72.73 & 52.54 & 69.55 & \secondplace{73.75} & \thirdplace{73.64} & 72.05 & 70.45 \\
census       & 5.92 & 14.47 & 11.12 & \thirdplace{20.25} & \secondplace{22.28} & 15.76 & 16.20 & 15.42 & \firstplace{40.44} & 15.93 & 17.96 \\
donors       & 86.54 & 83.34 & 45.17 & 26.65 & 52.09 & 95.54 & \secondplace{99.13} & 42.24 & 61.74 & \thirdplace{97.22} & \corefirst{99.96} \\
fault        & 50.67 & 55.57 & 54.41 & 52.24 & \thirdplace{57.50} & \secondplace{57.68} & 48.78 & 55.54 & \firstplace{58.13} & 54.86 & 55.30 \\
Hepatitis    & 35.29 & 30.77 & \firstplace{49.23} & \thirdplace{47.69} & 30.77 & 32.86 & \secondplace{48.69} & 35.38 & 44.62 & 40.00 & 41.54 \\
lympho       & 66.67 & 50.00 & 86.67 & \secondplace{96.67} & 0.00 & 86.67 & 77.78 & 83.33 & 53.33 & \thirdplace{90.00} & \corefirst{100.00} \\
mammography  & 41.15 & 40.77 & 42.38 & \thirdplace{46.69} & 43.46 & 44.14 & 30.15 & 40.00 & 40.85 & \secondplace{52.00} & \corefirst{53.15} \\
mnist        & \thirdplace{74.71} & 70.00 & 52.89 & 55.77 & 68.86 & \firstplace{80.14} & 61.65 & \secondplace{79.69} & 31.37 & 74.23 & \corethird{75.17} \\
musk         & \firstplace{100.00} & \firstplace{100.00} & 69.77 & 85.98 & \firstplace{100.00} & \secondplace{98.98} & 96.30 & \thirdplace{98.97} & 67.42 & \firstplace{100.00} & \corefirst{100.00} \\
optdigits    & \firstplace{84.00} & 55.52 & 11.60 & 0.80 & 8.00 & 75.73 & 62.50 & 28.00 & \secondplace{82.45} & \thirdplace{82.27} & 79.65 \\
Parkinson    & \thirdplace{89.12} & 85.71 & 86.94 & 88.30 & \thirdplace{89.12} & 84.90 & \firstplace{92.57} & 88.16 & \secondplace{90.61} & 88.30 & 88.71 \\
pendigits    & 72.44 & 71.15 & 53.08 & 22.69 & 55.13 & 67.82 & 82.60 & 46.54 & \thirdplace{83.72} & \firstplace{93.08} & \coresecond{92.05} \\
pima         & 64.93 & 64.93 & \secondplace{69.10} & 65.60 & \thirdplace{69.03} & 67.31 & 68.86 & 68.36 & 67.24 & 67.09 & \corefirst{69.55} \\
satimage-2   & 81.69 & 90.14 & 89.26 & 78.59 & \firstplace{94.37} & \secondplace{92.96} & 66.45 & 82.82 & 56.06 & \thirdplace{92.11} & 89.42 \\
shuttle      & 97.52 & 97.18 & 96.42 & 95.89 & 96.55 & 97.93 & 67.20 & \thirdplace{98.17} & 97.81 & \secondplace{98.55} & \corefirst{98.69} \\
thyroid      & 52.69 & \thirdplace{75.27} & \secondplace{79.78} & 72.26 & 70.97 & 73.83 & 66.90 & 72.47 & \firstplace{80.22} & 73.98 & 69.89 \\
wbc          & \secondplace{76.19} & 71.43 & 70.48 & 69.52 & 66.67 & \thirdplace{74.55} & 70.06 & \firstplace{80.00} & 73.33 & 71.43 & 73.33 \\
WDBC         & 90.00 & 90.00 & 88.00 & 76.00 & 80.00 & 90.91 & 89.89 & \secondplace{94.00} & \thirdplace{92.00} & \firstplace{96.00} & 90.00 \\
WPBC         & 36.17 & 34.04 & 35.74 & \thirdplace{37.45} & 34.04 & 36.25 & \firstplace{57.51} & 35.32 & \secondplace{42.55} & 32.34 & 35.74 \\
yeast        & 48.72 & 46.15 & 44.58 & 47.06 & 48.13 & 46.69 & \firstplace{67.99} & \thirdplace{49.78} & \secondplace{50.37} & 48.21 & 48.99 \\
\midrule
\rowcolor{gray!15}
\multicolumn{12}{c}{\textbf{Out-of-Domain Target Datasets}} \\
\midrule
amazon       & 8.20 & 9.20 & 8.20 & 8.92 & 8.40 & 7.56 & \firstplace{18.11} & 9.00 & \secondplace{12.72} & 9.64 & \corethird{11.12} \\
backdoor     & 40.63 & 50.19 & 0.78 & 15.13 & \secondplace{85.18} & 74.78 & 80.72 & \firstplace{86.73} & 15.61 & \thirdplace{82.35} & 76.32 \\
campaign     & 30.39 & 41.12 & 44.06 & 43.98 & 47.82 & \firstplace{57.97} & 42.31 & \thirdplace{49.44} & 43.52 & 46.46 & \coresecond{50.33} \\
cover        & 20.35 & 7.97 & 8.38 & 3.76 & 4.44 & 13.59 & 50.51 & \firstplace{81.00} & \secondplace{74.68} & 57.61 & \corethird{74.09} \\
fraud        & 0.00 & 35.57 & 32.32 & 26.59 & 35.57 & \secondplace{57.69} & 41.98 & 26.91 & \firstplace{63.25} & \thirdplace{48.71} & 41.72 \\
glass        & 0.00 & 0.00 & 2.22 & 0.00 & 0.00 & 8.00 & \secondplace{23.57} & 6.67 & \firstplace{40.00} & \thirdplace{17.91} & 11.11 \\
ionosphere   & 70.12 & 84.13 & 72.38 & 80.63 & 89.68 & 90.24 & 88.89 & \firstplace{93.49} & \secondplace{92.38} & \thirdplace{90.32} & \coresecond{92.38} \\
SpamBase     & 73.97 & 73.85 & \thirdplace{80.11} & 75.71 & 79.33 & 72.51 & 76.43 & \secondplace{80.58} & 63.10 & \firstplace{81.80} & \coresecond{81.38} \\
speech       & 1.64 & 1.64 & 3.61 & 1.64 & 1.64 & 3.87 & \secondplace{8.76} & \thirdplace{7.21} & \firstplace{10.49} & 4.26 & 3.61 \\
vowels       & 36.00 & 26.00 & 14.00 & 11.60 & 28.00 & \secondplace{38.80} & \thirdplace{37.74} & 35.20 & \firstplace{53.20} & 31.20 & 35.60 \\
Wilt         & 16.73 & 8.56 & 2.02 & 3.97 & 1.95 & 13.54 & \thirdplace{36.85} & \secondplace{39.17} & 9.65 & 12.45 & \corefirst{57.35} \\
\midrule
Average      & 54.40 & 55.01 & 50.18 & 48.67 & 51.93 & 60.12 & \summarythird{61.21} & 59.21 & 58.67 & \summarysecond{63.52} & \corefirst{65.31} \\
Rank $\downarrow$ & 7.15 & 7.25 & 7.46 & 8.07 & 6.91 & 5.46 & 5.53 & 5.15 & \summarythird{4.79} & \summarysecond{4.35} & \corefirst{3.88} \\
\bottomrule
\end{tabular}
}
\caption{F1-Score (\%) comparison between baselines and \ourmethod. Highlighted are the results ranked \firstplace{first}, \secondplace{second}, and \thirdplace{third}; colored cells indicate the rank of \ourmethod. ``Rank'' indicates the average ranking over all datasets.}
\label{tab:f1_results}
\end{table*}

%% file: Appendices/figs/context/detail_ratio_auc.tex
\begin{figure*}[!t]
    \centering

    \subfigure[arrhythmia]{
        \includegraphics[width=0.15\textwidth]{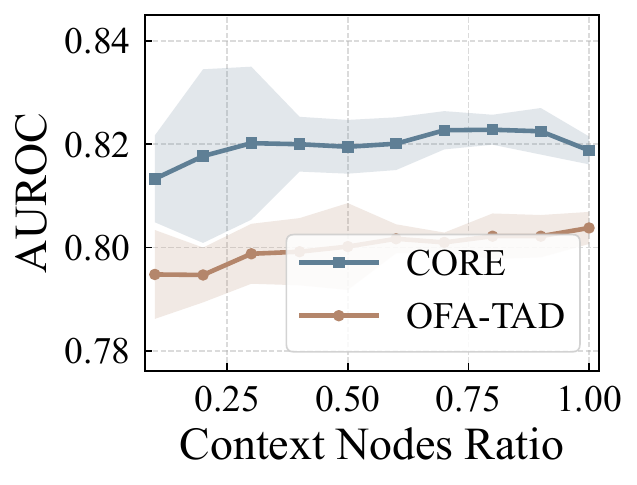}
    }
    \subfigure[mammography]{
        \includegraphics[width=0.15\textwidth]{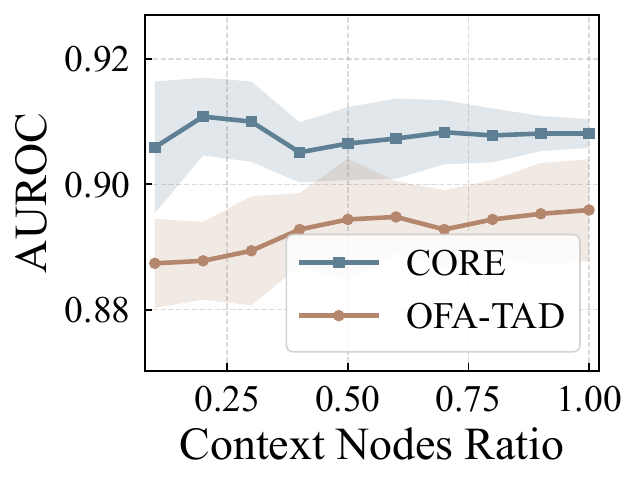}
    }
    \subfigure[backdoor]{
        \includegraphics[width=0.15\textwidth]{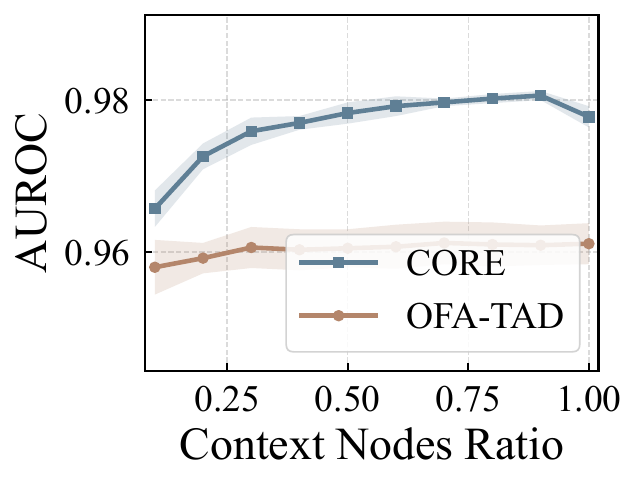}
    }
    \subfigure[ionosphere]{
        \includegraphics[width=0.15\textwidth]{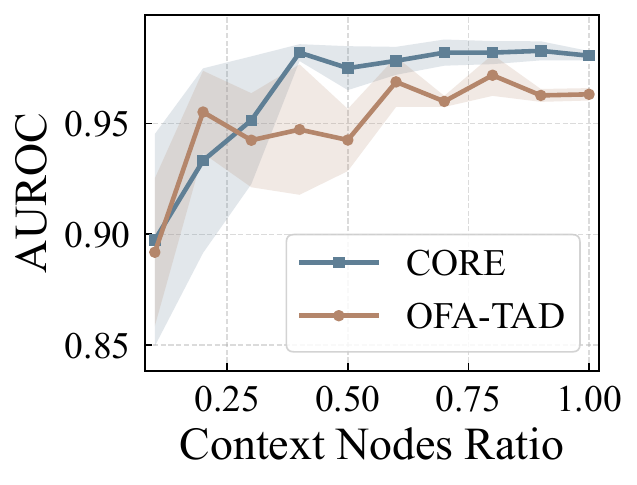}
    }
    \subfigure[Wilt]{
        \includegraphics[width=0.15\textwidth]{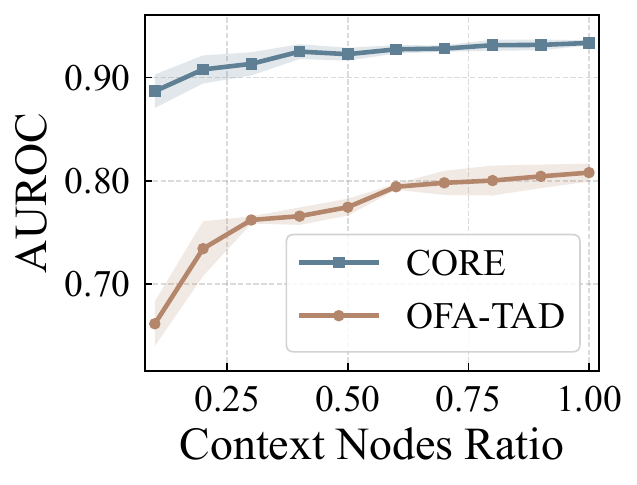}
    }
    \subfigure[Hepatitis]{
        \includegraphics[width=0.15\textwidth]{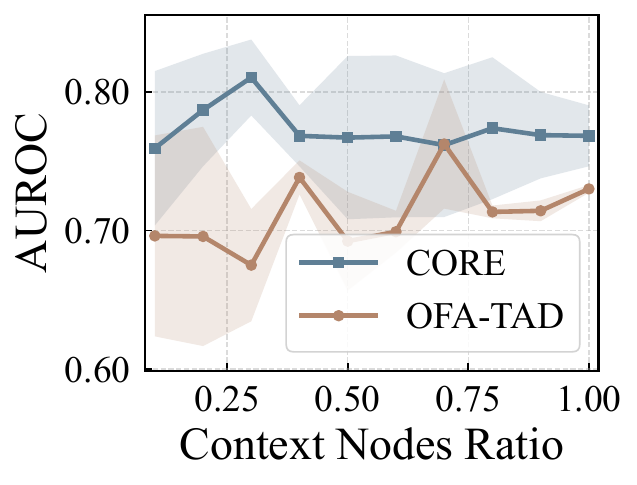}
    }
    
    \subfigure[Parkinson]{
        \includegraphics[width=0.15\textwidth]{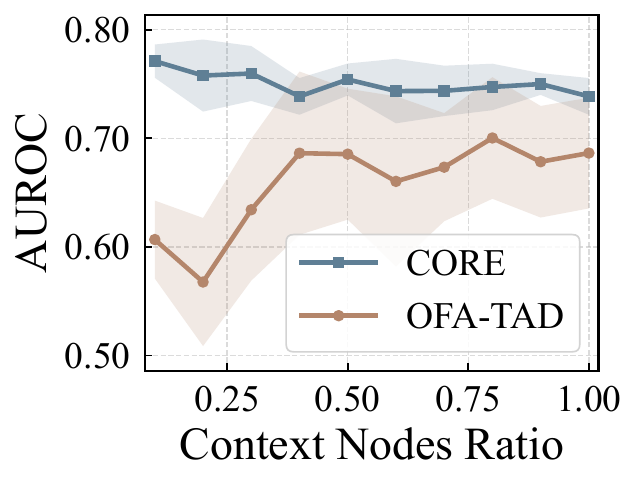}
    }
    \subfigure[WPBC]{
        \includegraphics[width=0.15\textwidth]{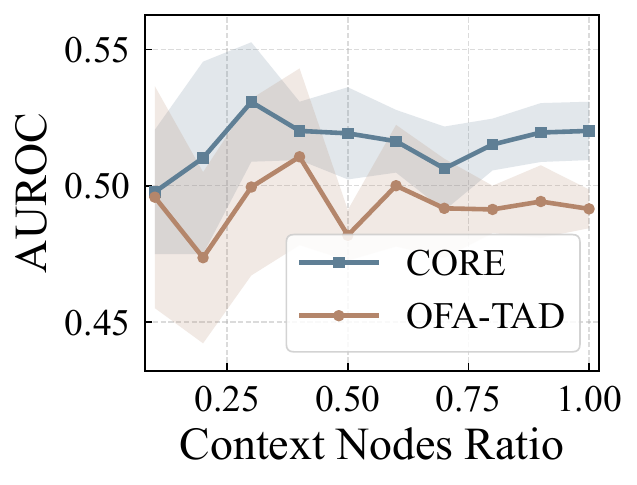}
    }
    \subfigure[campaign]{
        \includegraphics[width=0.15\textwidth]{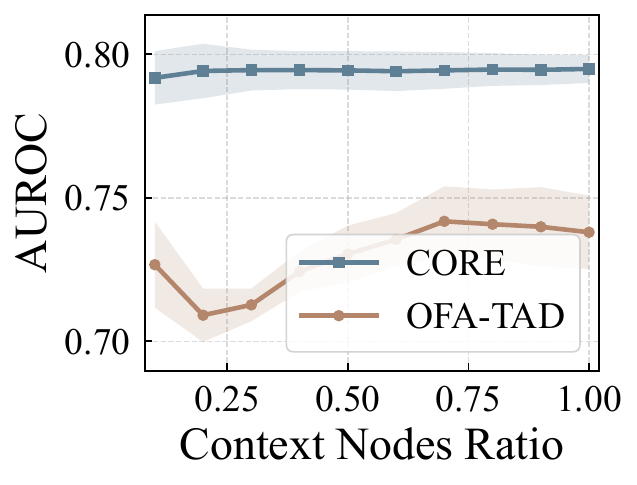}
    }
    \subfigure[glass]{
        \includegraphics[width=0.15\textwidth]{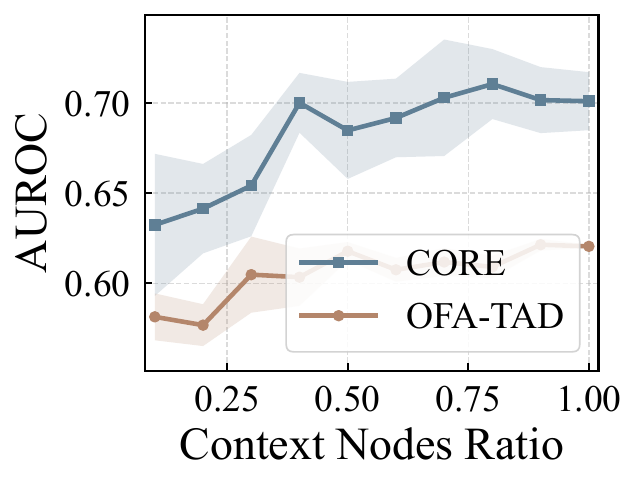}
    }
    \subfigure[cover]{
        \includegraphics[width=0.15\textwidth]{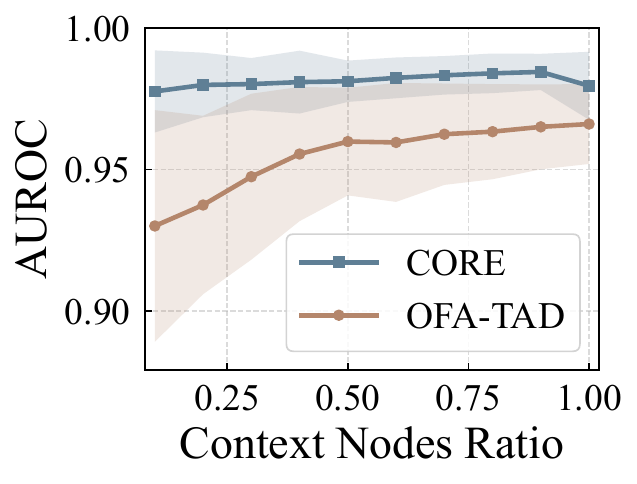}
    }
    \subfigure[fraud]{
        \includegraphics[width=0.15\textwidth]{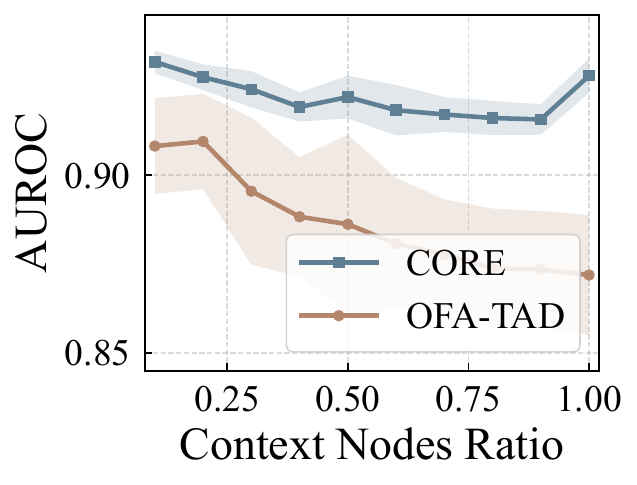}
    }

    \caption{AUROC performance with varying context nodes on twelve datasets.}
    \label{fig:context_ratio_auc_datasets}
\end{figure*}

%% file: Appendices/figs/context/detail_ratio_ap.tex
\begin{figure*}[!t]
    \centering

    \subfigure[arrhythmia]{
        \includegraphics[width=0.15\textwidth]{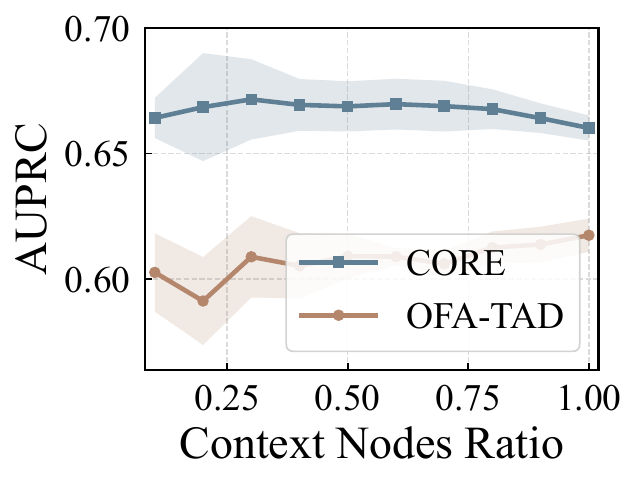}
    }
    \subfigure[mammography]{
        \includegraphics[width=0.15\textwidth]{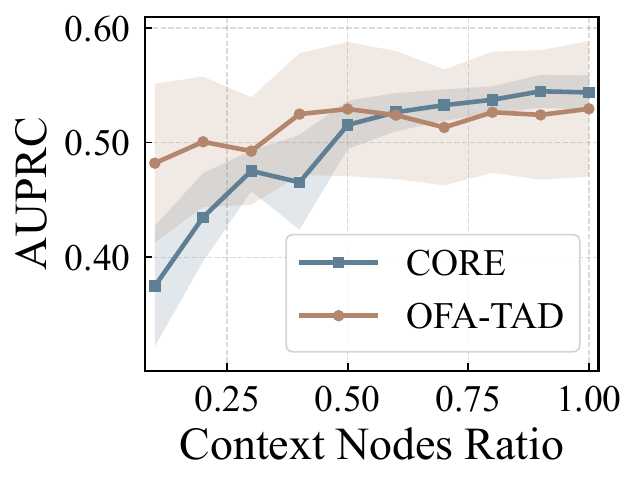}
    }
    \subfigure[backdoor]{
        \includegraphics[width=0.15\textwidth]{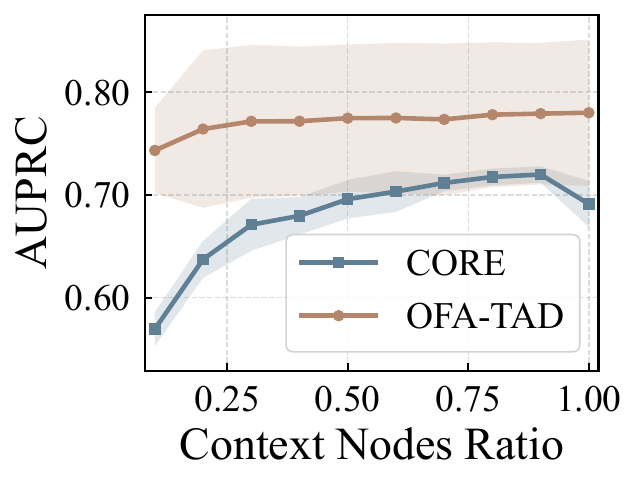}
    }
    \subfigure[ionosphere]{
        \includegraphics[width=0.15\textwidth]{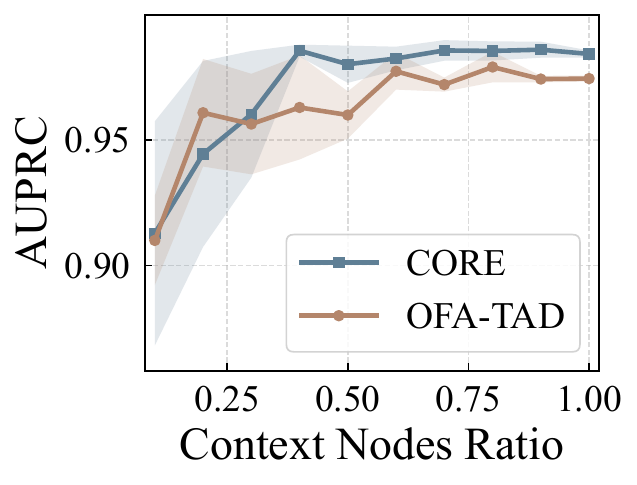}
    }
    \subfigure[Wilt]{
        \includegraphics[width=0.15\textwidth]{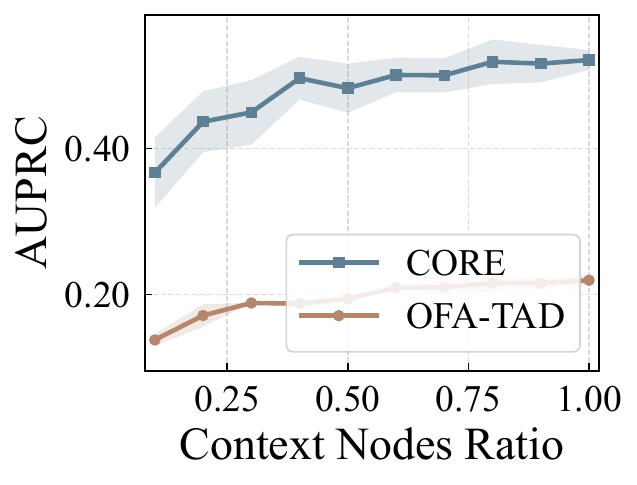}
    }
    \subfigure[Hepatitis]{
        \includegraphics[width=0.15\textwidth]{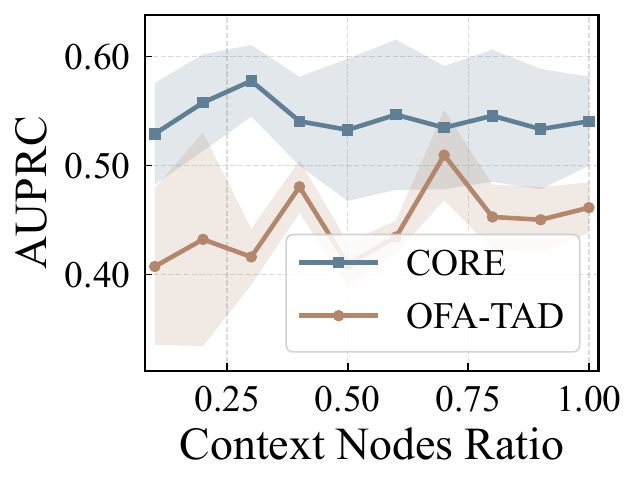}
    }
    
    \subfigure[Parkinson]{
        \includegraphics[width=0.15\textwidth]{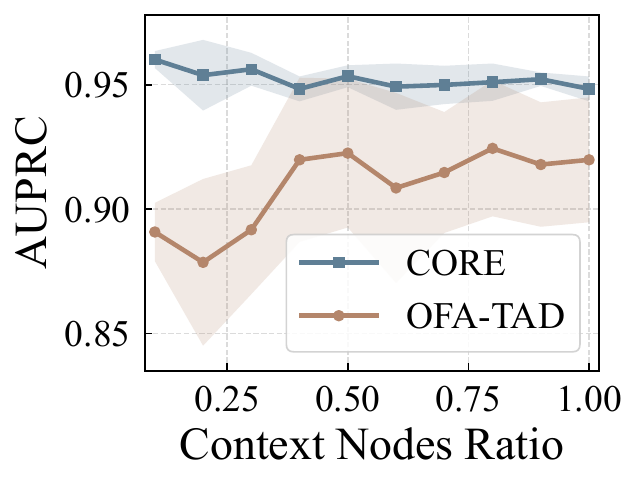}
    }
    \subfigure[WPBC]{
        \includegraphics[width=0.15\textwidth]{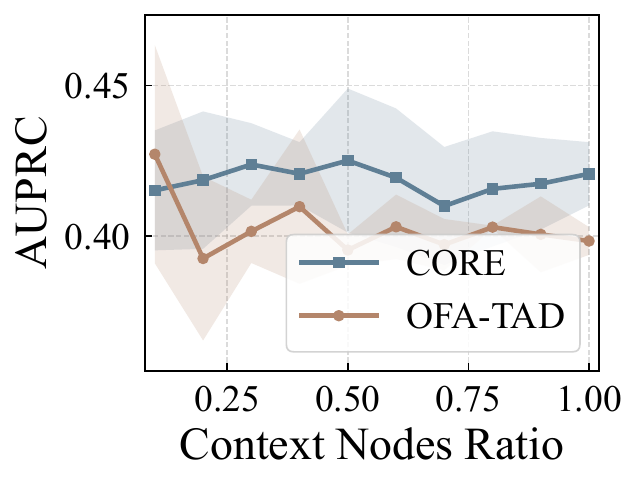}
    }
    \subfigure[campaign]{
        \includegraphics[width=0.15\textwidth]{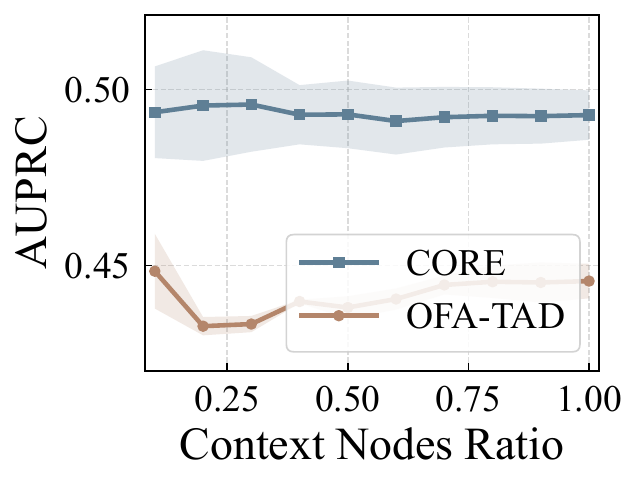}
    }
    \subfigure[glass]{
        \includegraphics[width=0.15\textwidth]{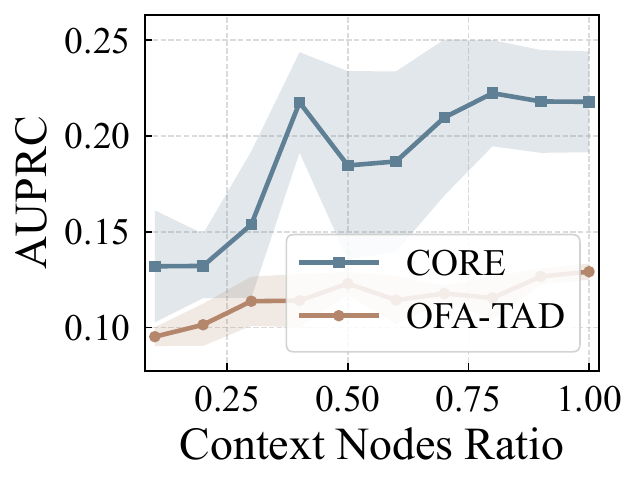}
    }
    \subfigure[cover]{
        \includegraphics[width=0.15\textwidth]{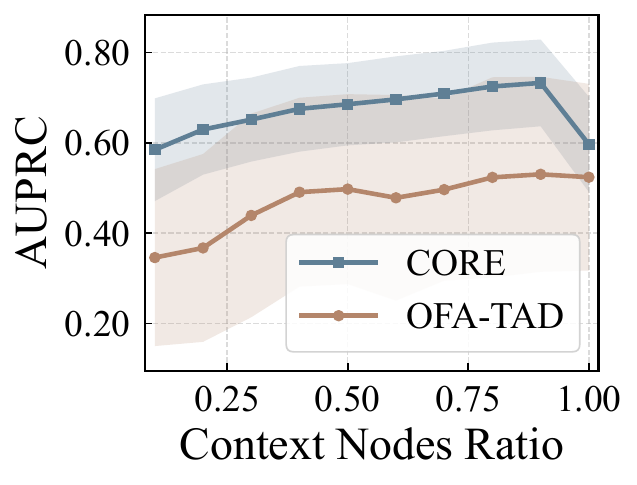}
    }
    \subfigure[fraud]{
        \includegraphics[width=0.15\textwidth]{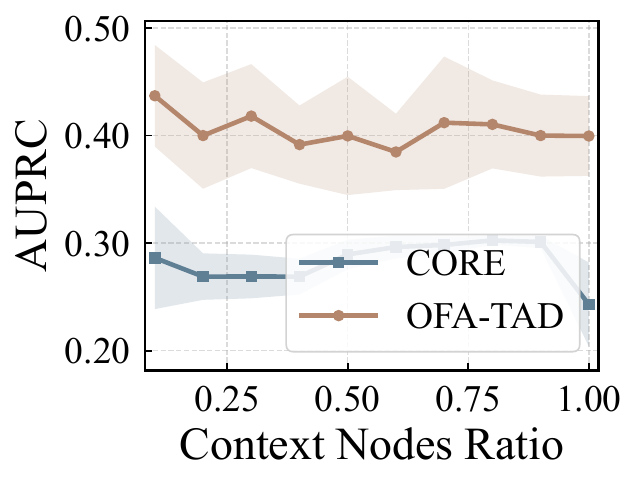}
    }

    \caption{AUPRC performance with varying context nodes on twelve datasets.}
    \label{fig:context_ratio_ap_datasets}
\end{figure*}

%% file: Appendices/figs/Efficiency/detail_ef.tex
\begin{figure*}[!t]
    \centering

    \subfigure[arrhythmia]{
        \includegraphics[width=0.31\textwidth]{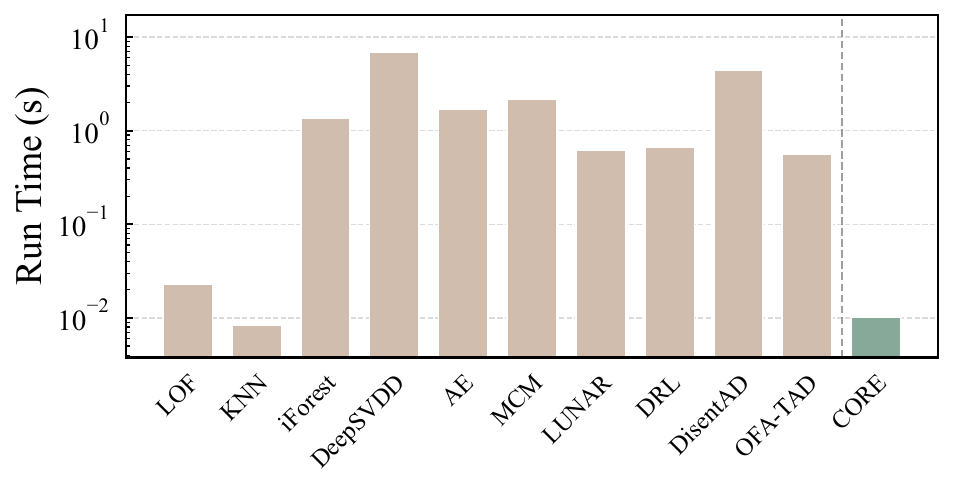}
    }
    \subfigure[mammography]{
        \includegraphics[width=0.31\textwidth]{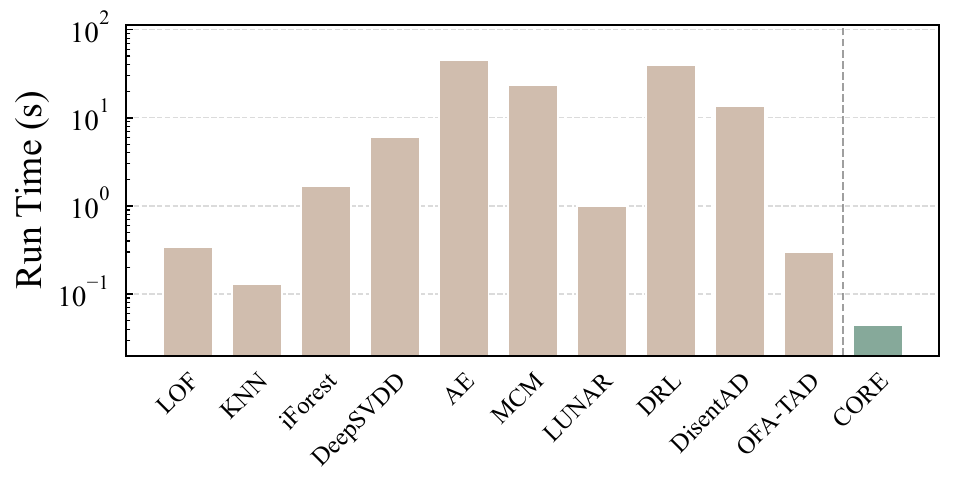}
    }
    \subfigure[backdoor]{
        \includegraphics[width=0.31\textwidth]{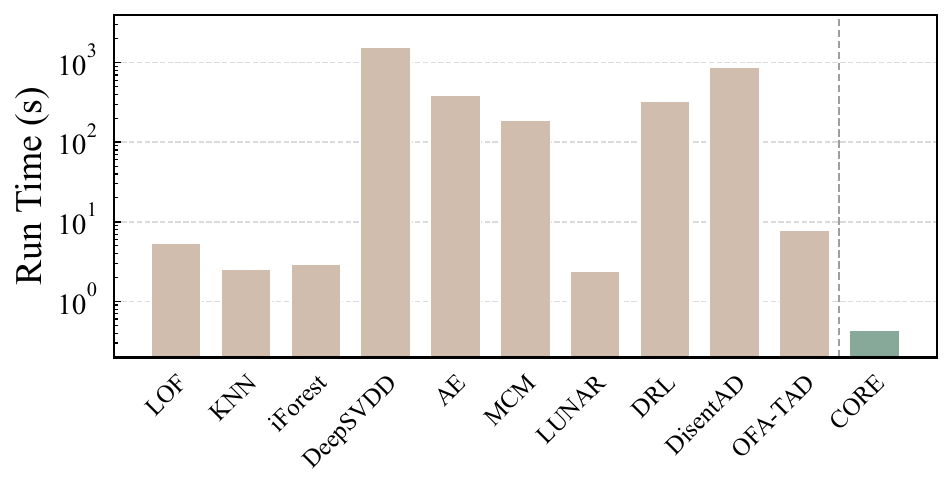}
    }

    \vspace{-0.6em}

    \subfigure[ionosphere]{
        \includegraphics[width=0.31\textwidth]{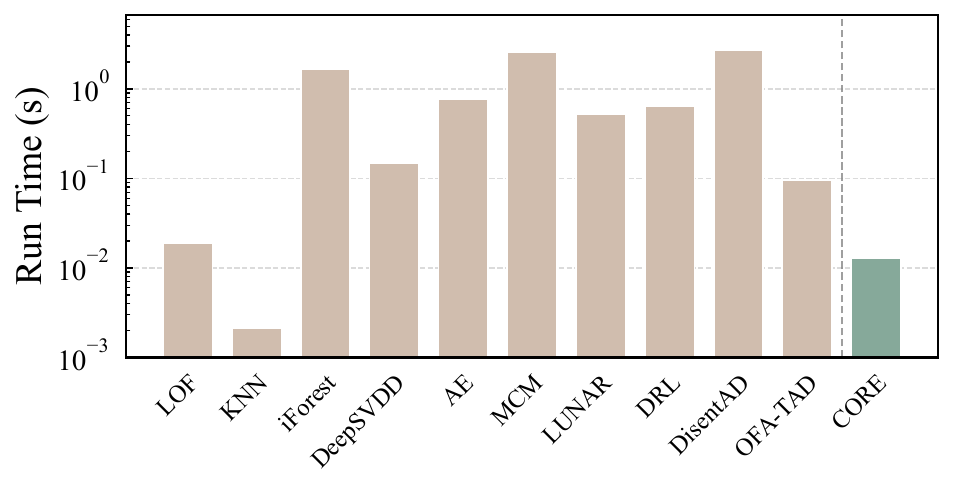}
    }
    \subfigure[Wilt]{
        \includegraphics[width=0.31\textwidth]{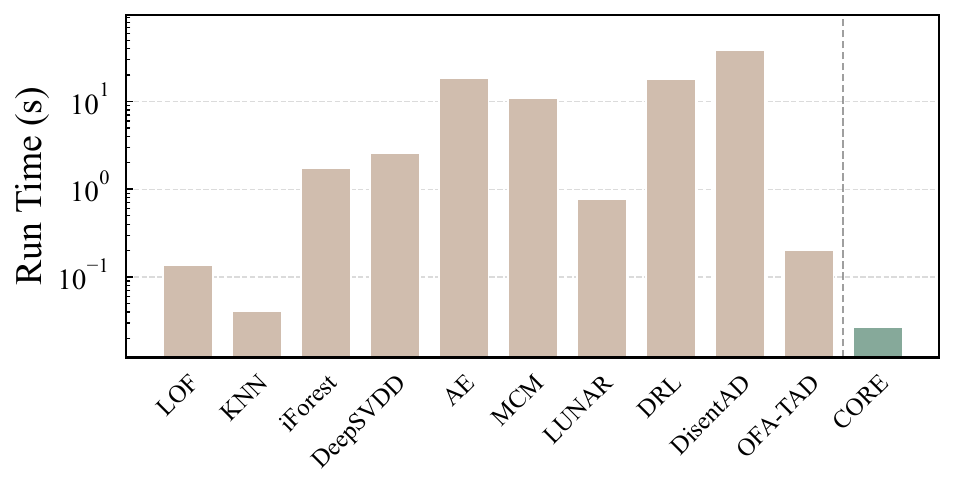}
    }
    \subfigure[Hepatitis]{
        \includegraphics[width=0.31\textwidth]{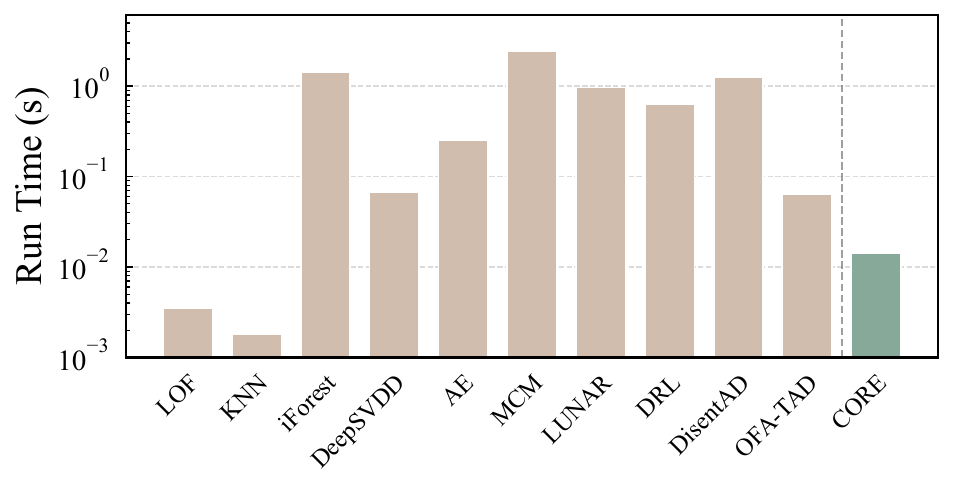}
    }

    \vspace{-0.6em}

    \subfigure[Parkinson]{
        \includegraphics[width=0.31\textwidth]{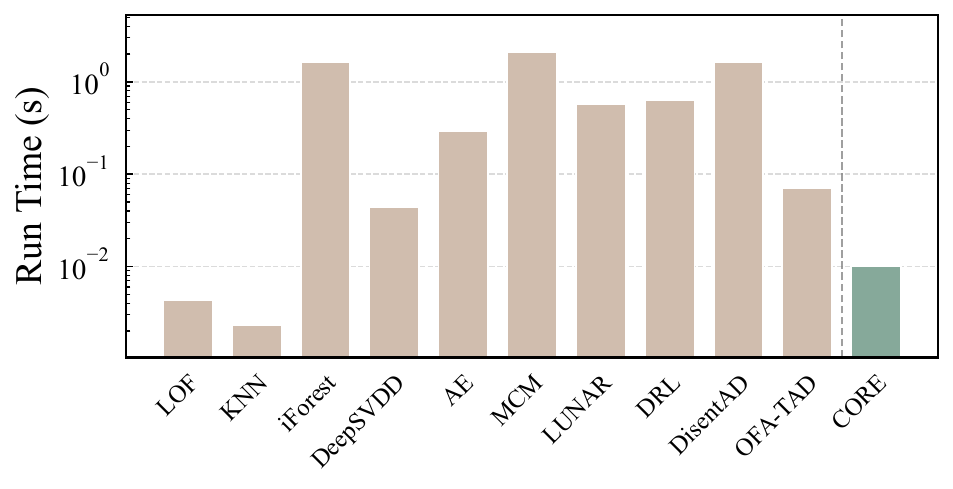}
    }
    \subfigure[WPBC]{
        \includegraphics[width=0.31\textwidth]{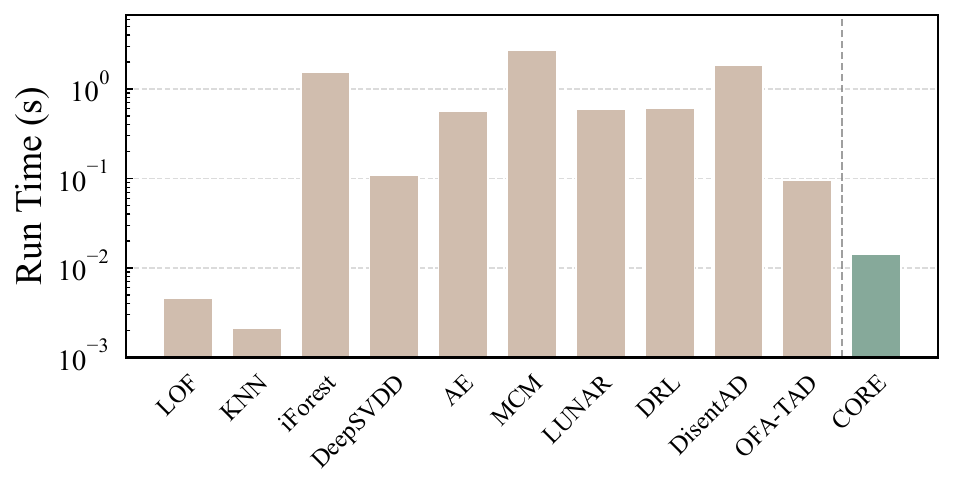}
    }
    \subfigure[campaign]{
        \includegraphics[width=0.31\textwidth]{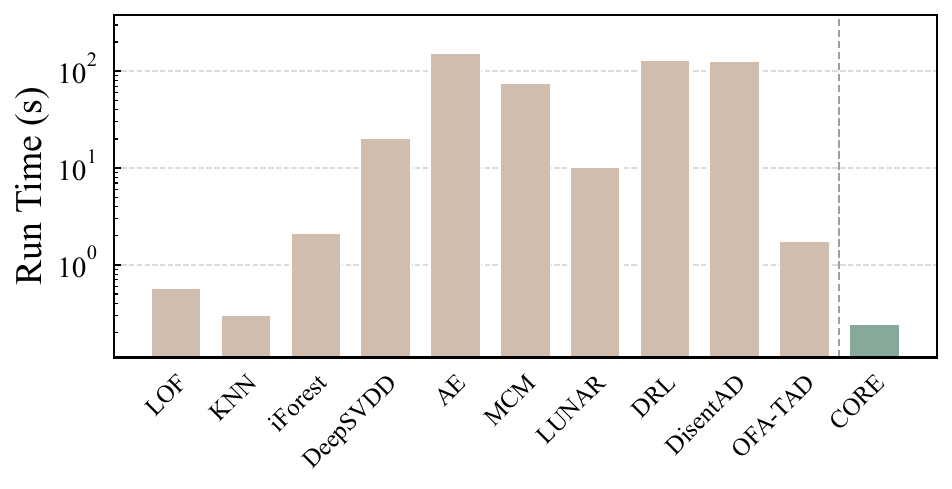}
    }

    \vspace{-0.6em}

    \subfigure[glass]{
        \includegraphics[width=0.31\textwidth]{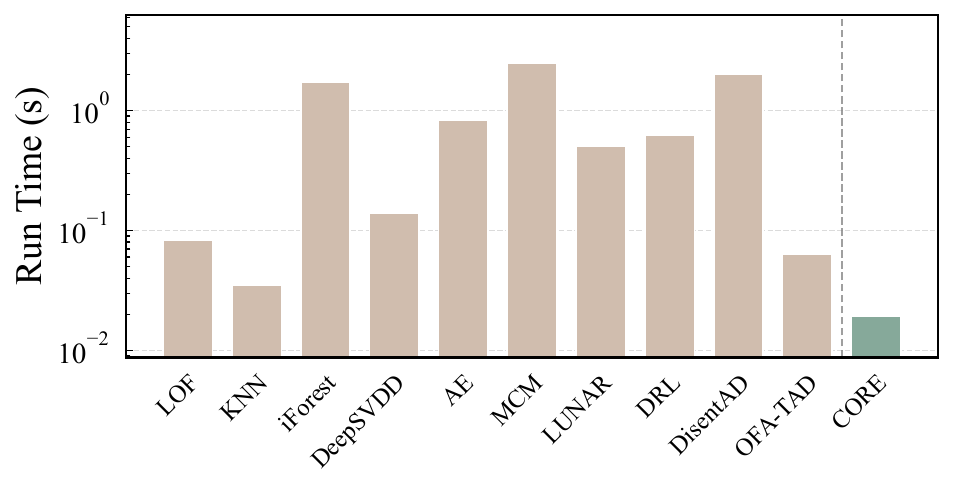}
    }
    \subfigure[cover]{
        \includegraphics[width=0.31\textwidth]{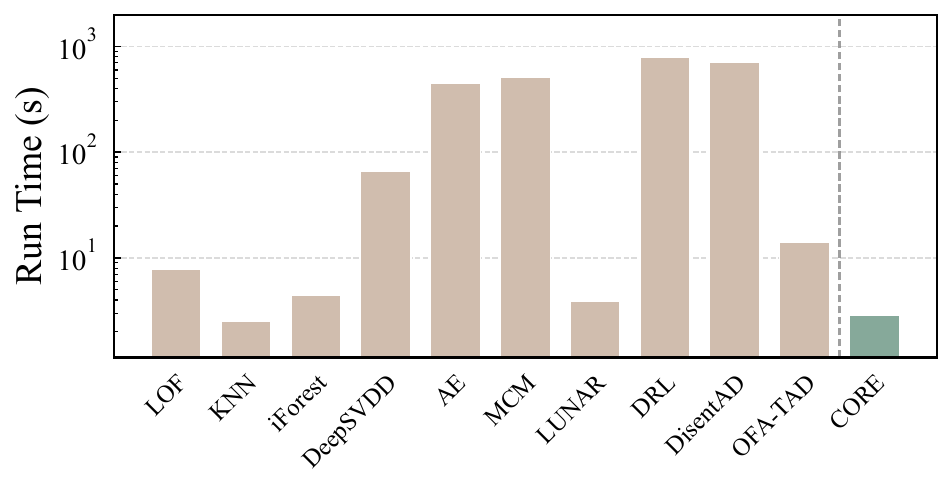}
    }
    \subfigure[fraud]{
        \includegraphics[width=0.31\textwidth]{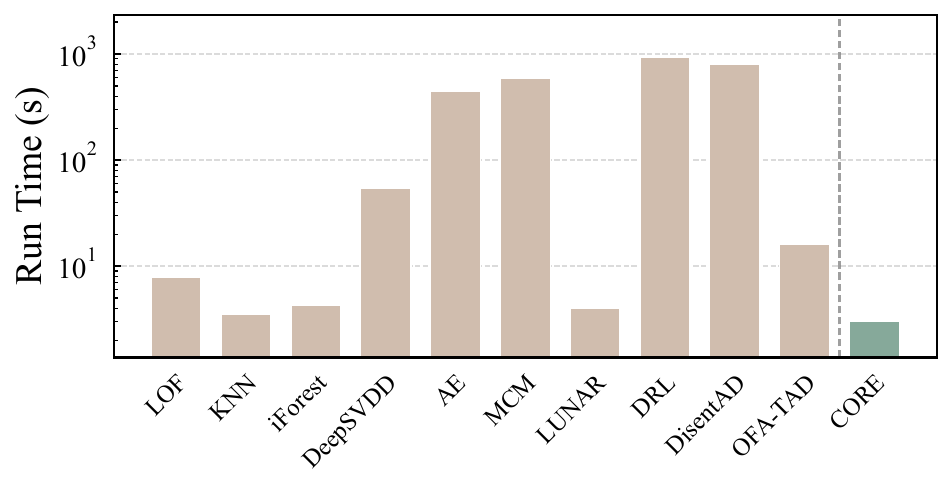}
    }

    \caption{Runtime comparison on twelve selected datasets.}
    \label{fig:ef_deatil}
\end{figure*}

%% file: aaai2027.bib
@article{fernando2021deep,
  title={Deep learning for medical anomaly detection--a survey},
  author={Fernando, Tharindu and Gammulle, Harshala and Denman, Simon and Sridharan, Sridha and Fookes, Clinton},
  journal={ACM computing surveys (CSUR)},
  volume={54},
  number={7},
  pages={1--37},
  year={2021},
  publisher={ACM New York, NY, USA}
}

@article{ahmad2021network,
  title={Network intrusion detection system: A systematic study of machine learning and deep learning approaches},
  author={Ahmad, Zeeshan and Shahid Khan, Adnan and Wai Shiang, Cheah and Abdullah, Johari and Ahmad, Farhan},
  journal={Transactions on Emerging Telecommunications Technologies},
  volume={32},
  number={1},
  pages={e4150},
  year={2021},
  publisher={Wiley Online Library}
}

@article{al2021financial,
  title={Financial fraud detection applying data mining techniques: A comprehensive review from 2009 to 2019},
  author={Al-Hashedi, Khaled Gubran and Magalingam, Pritheega},
  journal={Computer Science Review},
  volume={40},
  pages={100402},
  year={2021},
  publisher={Elsevier}
}

@inproceedings{yin2024mcm,
  title={Mcm: Masked cell modeling for anomaly detection in tabular data},
  author={Yin, Jiaxin and Qiao, Yuanyuan and Zhou, Zitang and Wang, Xiangchao and Yang, Jie},
  booktitle={The Twelfth International Conference on Learning Representations},
  year={2024}
}

@inproceedings{ye2025drl,
  title={DRL: Decomposed representation learning for tabular anomaly detection},
  author={Ye, Hangting and Zhao, He and Fan, Wei and Zhou, Mingyuan and Guo, Dandan and Chang, Yi},
  booktitle={International Conference on Learning Representations},
  volume={2025},
  pages={24554--24589},
  year={2025}
}

@inproceedings{goodge2022lunar,
  title={Lunar: Unifying local outlier detection methods via graph neural networks},
  author={Goodge, Adam and Hooi, Bryan and Ng, See-Kiong and Ng, Wee Siong},
  booktitle={Proceedings of the AAAI conference on artificial intelligence},
  volume={36},
  number={6},
  pages={6737--6745},
  year={2022}
}

@article{hollmann2025accurate,
  title={Accurate predictions on small data with a tabular foundation model},
  author={Hollmann, Noah and M{\"u}ller, Samuel and Purucker, Lennart and Krishnakumar, Arjun and K{\"o}rfer, Max and Hoo, Shi Bin and Schirrmeister, Robin Tibor and Hutter, Frank},
  journal={Nature},
  volume={637},
  number={8045},
  pages={319--326},
  year={2025},
  publisher={Nature Publishing Group UK London}
}

@article{ma2026tabdpt,
  title={TabDPT: Scaling Tabular Foundation Models on Real Data},
  author={Ma, Junwei and Thomas, Valentin and Hosseinzadeh, Rasa and Labach, Alex and Cresswell, Jesse and Golestan, Keyvan and Yu, Guangwei and Caterini, Anthony L and Volkovs, Maks},
  journal={Advances in Neural Information Processing Systems},
  volume={38},
  pages={172692--172722},
  year={2026}
}

@article{li2026towards,
  title={Towards One-for-All Anomaly Detection for Tabular Data},
  author={Li, Shiyuan and Liu, Yixin and Zheng, Yu and Cao, Xiaofeng and Pan, Shirui and Shen, Heng Tao},
  journal={arXiv preprint arXiv:2603.14407},
  year={2026}
}

@inproceedings{fu2025uniod,
  title={UniOD: A Universal Model for Outlier Detection across Diverse Domains},
  author={Fu, Dazhi and Fan, Jicong},
  journal={International Conference on Learning Representations},
  year={2026}
}

@inproceedings{ding2019deep,
  title={Deep anomaly detection on attributed networks},
  author={Ding, Kaize and Li, Jundong and Bhanushali, Rohit and Liu, Huan},
  booktitle={Proceedings of the 2019 SIAM international conference on data mining},
  pages={594--602},
  year={2019},
  organization={SIAM}
}

@article{zavrtanik2021reconstruction,
  title={Reconstruction by inpainting for visual anomaly detection},
  author={Zavrtanik, Vitjan and Kristan, Matej and Sko{\v{c}}aj, Danijel},
  journal={Pattern Recognition},
  volume={112},
  pages={107706},
  year={2021},
  publisher={Elsevier}
}

@inproceedings{liu2008isolation,
  title={Isolation forest},
  author={Liu, Fei Tony and Ting, Kai Ming and Zhou, Zhi-Hua},
  booktitle={2008 eighth ieee international conference on data mining},
  pages={413--422},
  year={2008},
  organization={IEEE}
}

@inproceedings{breunig2000lof,
  title={LOF: identifying density-based local outliers},
  author={Breunig, Markus M and Kriegel, Hans-Peter and Ng, Raymond T and Sander, J{\"o}rg},
  booktitle={Proceedings of the 2000 ACM SIGMOD international conference on Management of data},
  pages={93--104},
  year={2000}
}

@inproceedings{angiulli2002fast,
  title={Fast outlier detection in high dimensional spaces},
  author={Angiulli, Fabrizio and Pizzuti, Clara},
  booktitle={European conference on principles of data mining and knowledge discovery},
  pages={15--27},
  year={2002},
  organization={Springer}
}

@article{liznerski2020explainable,
  title={Explainable deep one-class classification},
  author={Liznerski, Philipp and Ruff, Lukas and Vandermeulen, Robert A and Franks, Billy Joe and Kloft, Marius and M{\"u}ller, Klaus-Robert},
  journal={arXiv preprint arXiv:2007.01760},
  year={2020}
}

@inproceedings{chen2018autoencoder,
  title={Autoencoder-based network anomaly detection},
  author={Chen, Zhaomin and Yeo, Chai Kiat and Lee, Bu Sung and Lau, Chiew Tong},
  booktitle={2018 Wireless telecommunications symposium (WTS)},
  pages={1--5},
  year={2018},
  organization={IEEE}
}

@inproceedings{ye2025disentangling,
  title={Disentangling tabular data towards better one-class anomaly detection},
  author={Ye, Jianan and Tan, Zhaorui and Hu, Yijie and Yang, Xi and Cheng, Guangliang and Huang, Kaizhu},
  booktitle={Proceedings of the AAAI Conference on Artificial Intelligence},
  volume={39},
  number={12},
  pages={13061--13068},
  year={2025}
}

@inproceedings{qiu2021neural,
  title={Neural transformation learning for deep anomaly detection beyond images},
  author={Qiu, Chen and Pfrommer, Timo and Kloft, Marius and Mandt, Stephan and Rudolph, Maja},
  booktitle={International conference on machine learning},
  pages={8703--8714},
  year={2021},
  organization={PMLR}
}

@inproceedings{shen2025fomo,
  title={FoMo-0D: A Foundation Model for Zero-shot Outlier Detection},
  author={Shen, Yuchen and Wen, Haomin and Akoglu, Leman},
  booktitle={1st ICML Workshop on Foundation Models for Structured Data},
  year={2025}
}

@article{ding2026zero,
  title={From Zero to Hero: Advancing Zero-Shot Foundation Models for Tabular Outlier Detection},
  author={Ding, Xueying and Wen, Haomin and Kl{\"u}tterman, Simon and Akoglu, Leman},
  journal={arXiv preprint arXiv:2602.03018},
  year={2026}
}

@article{kluttermann2026fomo,
  title={FoMo X: Modular Explainability Signals for Outlier Detection Foundation Models},
  author={Kl{\"u}ttermann, Simon and Katzke, Tim and Nguyen, Phuong Huong and M{\"u}ller, Emmanuel},
  journal={arXiv preprint arXiv:2603.17570},
  year={2026}
}

@article{wei2026iclad,
  title={ICLAD: In-Context Learning for Unified Tabular Anomaly Detection Across Supervision Regimes},
  author={Wei, Jack Yi and Armanfard, Narges},
  journal={arXiv preprint arXiv:2603.19497},
  year={2026}
}

@article{qian2026dynhd,
  title={DynHD: Hallucination Detection for Diffusion Large Language Models via Denoising Dynamics Deviation Learning},
  author={Qian, Yanyu and Tan, Yue and Liu, Yixin and Yu, Wang and Pan, Shirui},
  journal={arXiv preprint arXiv:2603.16459},
  year={2026}
}

@inproceedings{chen2025multi,
  title={Multi-Stage Verification-Centric Framework for Mitigating Hallucination in Multi-Modal RAG},
  author={Chen, Baiyu and Wongso, Wilson and Hu, Xiaoqian and Tan, Yue and Salim, Flora D},
  booktitle={KDD Cup Workshop for Multimodal Retrieval Augmented Generation},
  year={2025}
}

@article{shen2026raising,
  title={Raising the bar in graph ood generalization: Invariant learning beyond explicit environment modeling},
  author={Shen, Xu and Liu, Yixin and Wang, Yili and Miao, Rui and Dai, Yiwei and Pan, Shirui and Chang, Yi and Wang, Xin},
  journal={IEEE Transactions on Pattern Analysis and Machine Intelligence},
  year={2026}
}

@article{liu2026few,
  title={From Few-Shot to Zero-Shot: Towards Generalist Graph Anomaly Detection},
  author={Liu, Yixin and Li, Shiyuan and Zheng, Yu and Chen, Qingfeng and Zhang, Chengqi and Yu, Philip S and Pan, Shirui},
  journal={IEEE Transactions on Knowledge and Data Engineering},
  year={2026}
}

@inproceedings{liu2026rethinking,
  title={Rethinking Feature Alignment in Generalist Graph Anomaly Detection: A Relational Fingerprint-based Approach},
  author={Liu, Yujing and Liu, Yixin and Zheng, Yu and Liew, Alan Wee-Chung and Cao, Xiaofeng and Pan, Shirui},
  booktitle={International Conference on Machine Learning},
  year={2026}
}

@inproceedings{pan2026correcting,
  title={Correcting False Alarms from Unseen: Adapting Graph Anomaly Detectors at Test Time},
  author={Pan, Junjun and Liu, Yixin and Zhou, Chuan and Xiong, Fei and Liew, Alan Wee-Chung and Pan, Shirui},
  booktitle={Proceedings of the AAAI Conference on Artificial Intelligence},
  year={2026}
}

@article{zheng2026unsupervised,
  title={From unsupervised to few-shot graph anomaly detection: A multi-scale contrastive learning approach},
  author={Zheng, Yu and Jin, Ming and Liu, Yixin and Chi, Lianhua and Phan, Khoa T and Chen, Yi-Ping Phoebe},
  journal={Transactions on Graph Intelligence and Network Applications (TGINA)},
  year={2026}
}

@inproceedings{li2026ofa,
  title={{OFA-MAS}: One-for-All Multi-Agent System Topology Design based on Mixture-of-Experts Graph Generative Models},
  author={Li, Shiyuan and Liu, Yixin and Zheng, Yu and Li, Mei and Nguyen, Quoc Viet Hung and Pan, Shirui},
  booktitle={Proceedings of the ACM Web Conference},
  year={2026}
}

@article{li2026relational,
  title={Towards Anomaly Detection on Relational Data},
  author={Li, Shiyuan and Zhao, Yunfeng and Tan, Yue and Chen, Qingfeng and Liu, Yixin and Pan, Shirui},
  journal={arXiv preprint arXiv:2606.18621},
  year={2026}
}

@inproceedings{pan2026camera,
  title={CAMERA: Adapting to Semantic Camouflage in Unsupervised Text-Attributed Graph Fraud Detection},
  author={Pan, Junjun and Liu, Yixin and Zheng, Yu and Chi, Lianhua and Liew, Alan Wee-Chung and Pan, Shirui},
  booktitle={International Joint Conference on Artificial Intelligence},
  year={2026}
}

@article{tan2024influence,
  title={Influence-oriented personalized federated learning},
  author={Tan, Yue and Long, Guodong and Jiang, Jing and Zhang, Chengqi},
  journal={arXiv preprint arXiv:2410.03315},
  year={2024}
}

@inproceedings{zhao2026fedcigar,
  title={FedCIGAR: A Personalized Reconstruction Approach for Federated Graph-level Anomaly Detection},
  author={Zhao, Yunfeng and Liu, Yixin and Chen, Qingfeng and Li, Shiyuan and Tan, Yue and Pan, Shirui},
  booktitle={International Joint Conference on Artificial Intelligence},
  year={2026}
}

@article{borisov2022deep,
  title={Deep neural networks and tabular data: A survey},
  author={Borisov, Vadim and Leemann, Tobias and Se{\ss}ler, Kathrin and Haug, Johannes and Pawelczyk, Martin and Kasneci, Gjergji},
  journal={IEEE transactions on neural networks and learning systems},
  volume={35},
  number={6},
  pages={7499--7519},
  year={2022},
  publisher={IEEE}
}

@inproceedings{shenkar2022anomaly,
  title={Anomaly detection for tabular data with internal contrastive learning},
  author={Shenkar, Tom and Wolf, Lior},
  booktitle={International conference on learning representations},
  year={2022}
}

@article{thimonier2023beyond,
  title={Beyond individual input for deep anomaly detection on tabular data},
  author={Thimonier, Hugo and Popineau, Fabrice and Rimmel, Arpad and Doan, Bich-Li{\^e}n},
  journal={arXiv preprint arXiv:2305.15121},
  year={2023}
}

@article{pang2021deep,
  title={Deep learning for anomaly detection: A review},
  author={Pang, Guansong and Shen, Chunhua and Cao, Longbing and Hengel, Anton Van Den},
  journal={ACM computing surveys (CSUR)},
  volume={54},
  number={2},
  pages={1--38},
  year={2021},
  publisher={ACM New York, NY, USA}
}

@inproceedings{kim2019rapp,
  title={Rapp: Novelty detection with reconstruction along projection pathway},
  author={Kim, Ki Hyun and Shim, Sangwoo and Lim, Yongsub and Jeon, Jongseob and Choi, Jeongwoo and Kim, Byungchan and Yoon, Andre S},
  booktitle={International conference on learning representations},
  year={2019}
}

@inproceedings{livernoche2024diffusion,
  title={On diffusion modeling for anomaly detection},
  author={Livernoche, Victor and Jain, Vineet and Hezaveh, Yashar and Ravanbakhsh, Siamak},
  booktitle={International Conference on Learning Representations},
  volume={2024},
  pages={25836--25866},
  year={2024}
}

@article{parzen1962estimation,
  title={On estimation of a probability density function and mode},
  author={Parzen, Emanuel},
  journal={The annals of mathematical statistics},
  volume={33},
  number={3},
  pages={1065--1076},
  year={1962},
  publisher={JSTOR}
}

@article{kim2023odim,
  title={ODIM: Outlier detection via likelihood of under-fitted generative models},
  author={Kim, Dongha and Hwang, Jaesung and Lee, Jongjin and Kim, Kunwoong and Kim, Yongdai},
  journal={arXiv preprint arXiv:2301.04257},
  year={2023}
}

@article{ruff2021unifying,
  title={A unifying review of deep and shallow anomaly detection},
  author={Ruff, Lukas and Kauffmann, Jacob R and Vandermeulen, Robert A and Montavon, Gr{\'e}goire and Samek, Wojciech and Kloft, Marius and Dietterich, Thomas G and M{\"u}ller, Klaus-Robert},
  journal={Proceedings of the IEEE},
  volume={109},
  number={5},
  pages={756--795},
  year={2021},
  publisher={IEEE}
}

@article{golan2018deep,
  title={Deep anomaly detection using geometric transformations},
  author={Golan, Izhak and El-Yaniv, Ran},
  journal={Advances in neural information processing systems},
  volume={31},
  year={2018}
}

@article{brown2020language,
  title={Language models are few-shot learners},
  author={Brown, Tom and Mann, Benjamin and Ryder, Nick and Subbiah, Melanie and Kaplan, Jared D and Dhariwal, Prafulla and Neelakantan, Arvind and Shyam, Pranav and Sastry, Girish and Askell, Amanda and others},
  journal={Advances in neural information processing systems},
  volume={33},
  pages={1877--1901},
  year={2020}
}

@inproceedings{devlin2019bert,
  title={Bert: Pre-training of deep bidirectional transformers for language understanding},
  author={Devlin, Jacob and Chang, Ming-Wei and Lee, Kenton and Toutanova, Kristina},
  booktitle={Proceedings of the 2019 conference of the North American chapter of the association for computational linguistics: human language technologies, volume 1 (long and short papers)},
  pages={4171--4186},
  year={2019}
}

@article{pimentel2014review,
  title={A review of novelty detection},
  author={Pimentel, Marco AF and Clifton, David A and Clifton, Lei and Tarassenko, Lionel},
  journal={Signal processing},
  volume={99},
  pages={215--249},
  year={2014},
  publisher={Elsevier}
}

@article{scholkopf2001estimating,
  title={Estimating the support of a high-dimensional distribution},
  author={Sch{\"o}lkopf, Bernhard and Platt, John C and Shawe-Taylor, John and Smola, Alex J and Williamson, Robert C},
  journal={Neural computation},
  volume={13},
  number={7},
  pages={1443--1471},
  year={2001},
  publisher={MIT Press}
}

@article{li2022ecod,
  title={Ecod: Unsupervised outlier detection using empirical cumulative distribution functions},
  author={Li, Zheng and Zhao, Yue and Hu, Xiyang and Botta, Nicola and Ionescu, Cezar and Chen, George H},
  journal={IEEE Transactions on Knowledge and Data Engineering},
  volume={35},
  number={12},
  pages={12181--12193},
  year={2022},
  publisher={IEEE}
}
